\documentclass[10pt,twocolumn,letterpaper]{article}

\usepackage[pagenumbers]{cvpr} %

\usepackage[dvipsnames]{xcolor}

\usepackage{amsmath}
\usepackage{amssymb}
\usepackage{booktabs}
\usepackage{pifont}
\usepackage{lscape}
\DeclareMathOperator*{\argmin}{arg\,min}

\usepackage{tcolorbox}
\tcbuselibrary{theorems,skins}
\usepackage{multirow}
\usepackage{colortbl}

\definecolor{mygray}{rgb}{0.9,0.9,0.9}

\newcommand{\maxiou}{Max.IoU}
\newcommand{\ourmeasure}{LTSim}
\newcommand{\deepsim}{FID-FeatSim}
\newcommand{\mygood}{\textcolor{Green}{\ding{51}}}
\newcommand{\mybad}{\textcolor{Red}{\ding{53}}}
\newcommand{\mysoso}{\textcolor{Gray}{\ding{115}}}

\definecolor{cvprblue}{rgb}{0.21,0.49,0.74}
\usepackage[pagebackref,breaklinks,colorlinks,citecolor=cvprblue]{hyperref}

\usepackage[capitalize]{cleveref}
\crefname{section}{Section}{Sections}
\Crefname{section}{Section}{Sections}
\Crefname{table}{Table}{Tables}
\crefname{table}{Table}{Tables}

\def\paperID{****} %
\def\confName{CVPR}
\def\confYear{2024}

\title{\ourmeasure{}: Layout Transportation-based Similarity Measure \\ for Evaluating Layout Generation}

\author{Mayu Otani \qquad
Naoto Inoue \qquad
Kotaro Kikuchi \qquad
Riku Togashi \\
\textit{CyberAgent}
}

\begin{document}
\maketitle

\begin{abstract}
We introduce a layout similarity measure designed to evaluate the results of layout generation. 
While several similarity measures have been proposed in prior research, there has been a lack of comprehensive discussion about their behaviors.
Our research uncovers that the majority of these measures are unable to handle various layout differences, primarily due to their dependencies on strict element matching, that is one-by-one matching of elements within the same category. 
To overcome this limitation, we propose a new similarity measure based on optimal transport, which facilitates a more flexible matching of elements.
This approach allows us to quantify the similarity between any two layouts even those sharing no element categories, making our measure highly applicable to a wide range of layout generation tasks. 
For tasks such as unconditional layout generation, where FID is commonly used, we also extend our measure to deal with collection-level similarities between groups of layouts.
The empirical result suggests that our collection-level measure offers more reliable comparisons than existing ones like FID and \maxiou{}.
\end{abstract}

\section{Introduction}
Layout generation is the task of organizing elements, \eg, images and text in a document layout, to create a visually appealing composite.
This topic is primarily explored for graphic design applications, such as user interface development~\cite{rico} and magazine authoring~\cite{zheng19tog}.
Recent progress in generative models has led to the development of various models for this task
\cite{Kikuchi2021,li_layoutgan_2021,jiang_layoutformer_2023}.
Despite this growing interest in layout generation, the evaluation of generated layouts remains underexplored. 
Existing literature has introduced several automatic measures for assessing layout quality, yet it lacks a comprehensive discussion on their behaviors and limitations.
Our in-depth analyses reveal that we have overlooked these measures’ drawbacks.

\begin{figure}[t!]
    \centering
    \includegraphics[width=\linewidth]{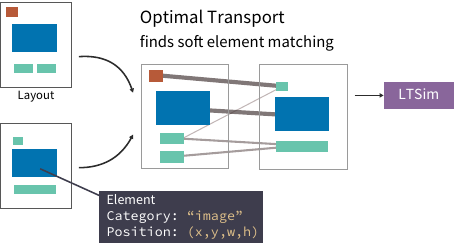}
    \caption{We propose \ourmeasure{}, a layout similarity measure based on optimal transportation. This flexible measure allows us to define the similarity among arbitrary layout pairs. Since \ourmeasure{} does not depend on learned representations, it can be applied to any dataset without the need for training.}
    \label{fig:teaser}
\end{figure}

In layout generation, a layout is represented by a set of elements.
Each element is a labeled bounding box whose label represents element categories, such as images, texts, or icons.
\Cref{fig:teaser} illustrates layout representations where the element color represents its category.
Evaluating the quality of generated layouts often involves comparing them to real ones, either individually or in groups.
For example, tasks like predicting element positions based on their categories are evaluated by comparing the generated layout with its real counterpart. 
For unconditional generation, we measure the discrepancy between real and generated distributions, similar to natural image generation evaluation.
Popular evaluation tools, such as DocSim~\cite{Patil_2020_CVPR_Workshops} and \maxiou{}~\cite{Kikuchi2021}, build element matching across layouts and aggregating the element-to-element similarities to derive a final score.

\begin{figure*}[t!]
    \centering
    \includegraphics[clip,width=\linewidth]{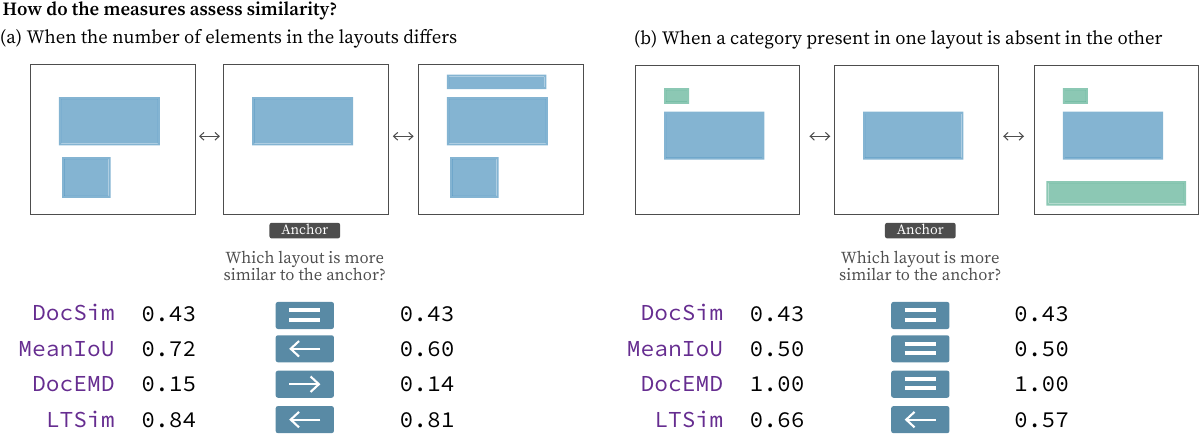}
    \caption{Existing measures are unable to quantify certain differences between layouts. (a) DocSim and DocEMD fail to identify which layout has smaller differences from the anchor. (b) All measures except ours judge that two layouts have the same similarity to an anchor.}
    \label{fig:examples}
\end{figure*}
One drawback of these similarity measures is that they deal with limited types of layout differences. 
As a result, they often fail to identify the most similar layout among differing ones. 
As shown in \Cref{fig:examples} (a) and (b), these measures struggle with layouts having different element numbers and category mismatches.
For instance, as seen in (a), DocSim gives the same similarity score to both layouts, while the right one has more differences from the anchor.
In (b), all existing measures fail to identify a more similar layout among two different ones. 
These issues arise from the presumption of specific types of differences, \ie, differences between elements within the same category. 
For instance, DocSim first finds one-by-one matches of elements in the same category and computes the similarity of matched elements, ignoring unmatched elements. 
Moreover, the existing measures do not consider cross-category element matching, imposing a constant penalty for missing categories.

To alleviate the issue, we propose a layout similarity measure, named \ourmeasure{} (\cref{fig:teaser}).
We design the optimal transportation-based similarity, inspired by the work for object detection evaluation~\cite{Otani_2022_CVPR}.
Optimal transportation allows for flexible element matching, including many-to-many and cross-category matching.
This flexibility expands the application of similarity-based evaluation from limited to arbitrary layout pairs such as those having extra elements and category mismatches.
We further extend our measure to a collection-level measure using Maximum Mean Discrepancy (MMD)~\cite{MMD_gretton12a}, enabling comparison between two layout groups. 
While FID~\cite{fid_neurips_2017} is commonly used for such comparisons as in image generation, our measure offers a distinct advantage; it eliminates the need for a dataset-specific feature extractor. 
This broadens our measure's applicability across various domains including UI and document layouts, without training.

Our contributions are summarized as follows:
\begin{itemize}
\setlength{\parskip}{0cm}
\setlength{\itemsep}{1mm}
    \item We develop a similarity measure, \ourmeasure{}, based on optimal transport over layout elements. Our measure quantifies layout differences that are overlooked by existing measures.
    \item We extend the layout similarity for collection-level comparison. Unlike FID, this measure does not require representation learning for specific datasets.
\end{itemize}
While realizing these, our measure follows basic similarity criteria of existing ones, ensuring no catastrophic gaps from results in prior research.

\section{Related Work}
\subsection{Layout Generation}
The task of layout generation is actively addressed in the computer graphics and vision communities. Earlier work adapts an optimization-based approach that constructs a cost function to measure layout quality and optimizes the layout parameters~\cite{odonovan_learning_2014,odonovan_designscape_2015}. Recent studies focus on learning a layout generation model~\cite{huang_survey_2023} and consider real-world layouts as a proxy of high-quality layouts, rather than manually defining the layout quality. The generation models include the generative models, such as VAE~\cite{Arroyo_2021_CVPR,Patil_2020_CVPR_Workshops,lee_neural_2020}, GAN~\cite{li_layoutgan_2021,Kikuchi2021}, diffusion models~\cite{play_icml23,inoue2023layout,docemd}, as well as those based on maximum likelihood estimation~\cite{gupta_layouttransformer_2021,kong_blt_2022,jiang_layoutformer_2023}.

In practice, the user of a model often desires to control the model's output to satisfy user-specific requirements. For this reason, two different problem settings are explored: \emph{unconditional generation}, generating layouts from scratch, and \emph{conditional generation}, generating layouts from partial or noisy observations, such that the element categories are known, but the positions and sizes are unknown.

\begin{table*}[t!]
\caption{Layout similarity measures. `Sample Equality' represents whether the measure has an equal scale across layouts. `Sensitivity \#Elements/Categories' shows if the measure can quantify the extent of the differences. \mysoso under `Computational Cost' indicates that advanced computing resources, like GPUs or parallel computing, may be needed to speed up the process. `Uses' shows examples of papers that use the measure for evaluation. For a comprehensive list of the usage by recent papers, please refer to the supplementary material.}
\centering
\begin{tabular}{@{}rccccccl@{}}
\toprule
                                   & \shortstack{Comp.\\Level} & \shortstack{Samp.\\Equality}                       &  \shortstack{Sens.\\\#Elem.}&\shortstack{Sens.\\Categ.}& \shortstack{Train-\\free}          & \shortstack{Comput.\\Cost}      &Uses\\ \midrule
DocSim                             & layout     & \mybad          & \mybad&\mybad& \mygood & \mygood  &\cite{kong_blt_2022,inoue2023layout}\\
MeanIoU                           & layout     & \mygood         & \mygood  &\mybad& \mygood & \mygood  &\cite{manandhar_eccv22,BaiMWC023}\\
DocEMD                             & layout     & \mybad          & \mygood  &\mybad& \mygood & \mygood   &\cite{docemd}\\
\ourmeasure{}                      & layout     & \mygood         & \mygood  &\mygood& \mygood & \mygood  &---\\ \midrule
\maxiou{}                          & collection   & ---          & \mybad   &\mybad& \mygood & \mygood  &\cite{Kikuchi2021,inoue2023layout,jiang_layoutformer_2023}\\
FID                                & collection   & ---             & \mygood  &\mygood  & \mybad  & \mysoso  &\cite{Kikuchi2021,inoue2023layout,jiang_layoutformer_2023,kong_blt_2022}\\
\ourmeasure{}-MMD                  & collection   & ---             & \mygood  &\mygood  & \mygood & \mysoso   &---\\ \bottomrule
\end{tabular}
    \label{tab:measure_summary}
\end{table*}

\subsection{Similarity Learning}
Prior works in~\cite{Patil_2021_CVPR,BaiMWC023,manandhar_eccv22} address representation learning or similarity learning for downstream tasks such as layout retrieval.
They introduce neural network models tailored for layouts and train the models on IoU-based similarity \cite{Patil_2021_CVPR,manandhar_eccv22} or reconstruction supervision \cite{BaiMWC023}.
While these studies predict layout similarity, they haven't been adopted for layout generation evaluation due to inherent challenges like the need for dataset-specific training and performance instability. 
These obstacles pose critical issues in the development of a reliable evaluation tool.

Our approach bypasses such issues by eliminating data-specific representation learning.
Our proposed measure is inspired by OC-cost, a dissimilarity measure to compare detection results~\cite{Otani_2022_CVPR}.
Oc-cost evaluates object detection results by assessing the cost of correcting predicted bounding boxes to ground-truth ones.
Our measure is also based on the cost of moving elements from one layout to the other.

\section{Review of Layout Generation Evaluation}
Previous studies have created tools to evaluate layout generation, yet their behaviors are rarely discussed. 
For the benefit of future research, we offer a comprehensive review of layout generation evaluation.
The characteristics of similarity measures are summarized in \cref{tab:measure_summary}.
We classify measures into layout-level and collection-level comparisons. 
The former compares two layouts, while the latter compares two groups of layouts.

In this paper, we represent a layout as a collection of $n$ elements $\mathcal{L}=\left\{ e_1, \ldots, e_n \right\}$.
Each element is denoted as $e_i = (b_i, c_i)$, where $b_i$ is a bounding box's normalized coordinates, and $c_i$ is a label indicating the element category. Let $\mathcal{C} = \{\mathcal{L}_1, \ldots, \mathcal{L}_s\}$ be a collection of real layouts, and $\hat{\mathcal{C}} = \{\hat{\mathcal{L}}_1, \ldots, \hat{\mathcal{L}}_t\}$ be generated ones.

\paragraph{DocSim}
DocSim is a similarity for a layout pair~\cite{Patil_2020_CVPR_Workshops}. 
In DocSim, each element in one layout is connected to an element in another. 
The weight of these connections shows how similar the elements are.
The weight is defined as:
\begin{equation}
    W(e_i, \hat{e}_j) = \alpha (b_i, \hat{b}_j)2^{-\Delta_{C}(b_i, \hat{b}_j)-\Delta_{S}(b_i, \hat{b}_j)},
\end{equation}
where $\alpha(\cdot, \cdot)$ is the smaller size of the element pair, $\Delta_{C}(\cdot, \cdot)$ is the relative Euclidean distance between the elements' center locations, and $\Delta_{S}(\cdot, \cdot)$, the shape difference. When an element pair has different categories, $W(e_i, e_j)=0$.
After calculating the weights for edges, the maximum weight matching is obtained, and the weights of the matching are averaged as a similarity score of the pair. 

\begin{figure}[t!]
    \centering
    \includegraphics[clip,width=0.9\linewidth]{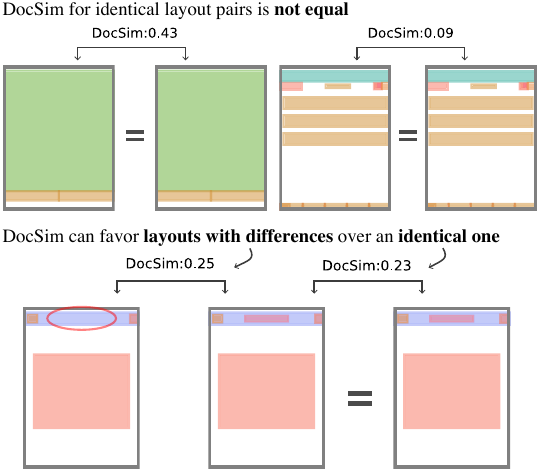}
    \caption{DocSim's drawbacks. Top: DocSim assigns much higher similarity values for layout pairs with larger elements than those with small elements. Bottom: DocSim happens to reward differences between layouts.}
    \label{fig:docsim_limitations}
\end{figure}

DocSim has the following drawbacks:
\begin{itemize}
    \item \textbf{Inconsistent scale}: The element size determines the weight scale with $\alpha(\cdot, \cdot)$. Due to this design, the upper bound of DocSim varies depending on the element size as in \cref{fig:docsim_limitations}. This means that layouts with larger elements have much more impact on evaluation scores.
    \item \textbf{Unexpected rewards for differences}: Since the DocSim score is the average weight of matched elements, removing small elements can result in a higher score than the one for an identical pair (\cref{fig:docsim_limitations}).
\end{itemize}

\paragraph{MeanIoU} %
MeanIoU is a common measure in segmentation tasks. 
MeanIoU calculates the intersection of two segmentation maps in the same category divided by their union.
This is done for each class and the average is taken.
Layout representation learning research~\cite{manandhar_eccv22,BaiMWC023} uses a MeanIoU-based measure for layout retrieval evaluation.
While MeanIoU is uncommon in layout generation research, it can be an option when viewing a layout as a labeled segmentation map.
One limitation is the constant penalty for the occurrence of categories that do not appear in one layout as in \cref{fig:examples}~(b).
Each category mismatch results in a zero IoU, causing a decrease in MeanIoU.

\paragraph{Fr\'{e}chet Inception Distance (FID)}
FID~\cite{fid_neurips_2017} is a common evaluation tool for data generation tasks. 
FID measures the discrepancy between the distributions of real and generated samples within a feature space. 
The performance of FID hence relies on a feature extractor. 
However, in contrast to image generation evaluation, there's no consensus in the layout generation community on the choice of a feature extractor.
Some researches~\cite{Kikuchi2021,inoue2023layout} directly encode elements' positions and labels by a Transformer-based model.
Conversely, others~\cite{play_icml23} render layouts to raster images and convert them into Inception features~\cite{inception}, as is done in image generation evaluations.

\paragraph{Maximum IoU (\maxiou{})}
\maxiou{}, proposed in~\cite{Kikuchi2021}, is a training-free, collection-level metric.
\maxiou{} was originally designed for label-conditional layout generation where layouts are generated from label multisets sampled from a reference collection.
Thus, this measure only evaluates layouts that have a corresponding one in the reference collection with the same label multiset.

The computation of \maxiou{} is also based on maximum weight matching. Let $\mathcal{P}^\ast=\{(\mathcal{L}_i, \hat{\mathcal{L}_i}) \mid i=1,\ldots,\min(|\mathcal{C}|, |\hat{\mathcal{C}}|)\}$ be a maximum weight matching between layout collections $\mathcal{C}$ and $\hat{\mathcal{C}}$, \maxiou{} is computed as:
\begin{align}
    \mathrm{MaxIoU}(\mathcal{C}, \hat{\mathcal{C}}) = \frac{1}{|\mathcal{P}^\ast|}\sum_{\mathcal{P}^\ast} \mathrm{MaxIoU}_{\mathrm{\beta}}(\mathcal{L}_i, \hat{\mathcal{L}}_i)
\end{align}
$\mathrm{MaxIoU}_{\mathrm{\beta}}(\cdot, \cdot)$ is a layout-level similarity that is also computed by obtaining maximum weight matching $\mathcal{Q}^\ast=\{(b_i, \hat{b}_i) \mid c_i = \hat{c}_i, i=1,\ldots,\min(|\mathcal{L}|, |\hat{\mathcal{L}}|)\}$ in a bipartite graph between two element groups.
\begin{equation}
    \mathrm{MaxIoU}_{\mathrm{\beta}}(\mathcal{L}_i, \hat{\mathcal{L}}_j) = \frac{1}{|\mathcal{Q}^\ast|}\sum_{\mathcal{Q}^\ast} \mathrm{IoU}(b_i, \hat{b}_i).
\end{equation}
IoU is Intersection over Union between two elements.

Unlike DocSim, a similarity score for a layout pair is normalized between 0 and 1, thus each sample equally contributes to the final performance score.
The limitation is that \maxiou{} evaluates only layout pairs with an identical label multiset.
This requirement can result in ignoring the majority of samples when elements' categories are also the generation target. 
While recent studies involving category prediction~\cite{jiang_layoutformer_2023,inoue2023layout} evaluate models using this measure, we reveal its unreliability in such a setting.

\paragraph{Layout Principles}
\textit{Overlap} and \textit{Alignment} are often considered to check for design rule violations~\cite{li_layoutgan_2021}.
The assumption behind this principle is that elements in a good layout should overlap less and align more. 
Overlap calculates the total overlapping area percentage among all element pairs. 
Alignment rewards six types of alignment in element pairs: horizontal/vertical center, top, bottom, left, and right alignments. 
As real layouts may not always adhere to these layout principles, we recommend reporting the scores of real layouts when evaluating a model using this measure. 
If a model scores better than real layouts, the model may be overly optimized for these layout principles.

\paragraph{Others} There are some relevant measures but their use is not common in layout generation.
\textbf{DocEMD}~\cite{docemd} is proposed for evaluating document layout creation.
It views elements as a 2D point set spread in their areas and measures layout dissimilarity using the summation of Earth Mover's Distance (EMD) on these sets.
Similar to DocSim, its scale varies by layout, leading to unstable impacts on evaluation scores.
DocEMD is also unable to quantify differences with category mismatches.
Our measure also uses EMD over two element sets but offers a more flexible cost for EMD so that it quantifies layout differences with category mismatches.
\textbf{Wasserstein distance}~\cite{Arroyo_2021_CVPR} is a collection-level similarity measure.
This measure views a layout collection as a distribution of element categories and element coordinates.
It respectively computes EMD for the category and coordinates distributions to assess the discrepancy between real and generated layouts.
Although this measure provides a rough perspective of the discrepancy, the oversimplified comparison may lead to unreliable conclusions.

We summarize challenges in layout similarities as below.
\begin{itemize}

\item \textbf{Restriction on element matching}: DocSim and \maxiou{} require one-to-one element matching,thereby limiting their application to tasks. 
MeanIoU and DocEMD mitigate this by grouping elements of the same category and representing them with 2D points or segmentation maps. 
However, these measures impose constant penalties for category mismatches and fail to quantify fine-grained differences between layouts.

\item \textbf{Need for dataset-specific feature extractor}: Layout generation applications span various fields~\cite{rico,publaynet,canvasvae}.
Evaluation measures like FID, which rely on a pre-trained feature extractor, may not be applicable in certain domains due to insufficient layout datasets for pre-training these extractors. 
Furthermore, prior work reports that FID does not always agree with perceived quality~\cite{yang2023layout,ding_cogview2_2022}.
\end{itemize}

\section{Layout Transportation Similarity}
This work offers two layout similarity measures for comparison at both layout and collection levels.

\subsection{Layout-level Similarity}
We define the dissimilarity between two layouts as the optimal cost to move elements from one layout to the other. 
the cost is obtained by solving this optimization problem:
\begin{align}
\gamma^\ast &= \argmin_{\gamma} \sum_{i,j} \gamma_{i,j} \mu(e_i, \hat{e}_j), \\
\text{s.t.} \, \sum_{i=1}^m\gamma_{i,j} &= \frac{1}{n}, \, \sum_{j=1}^n\gamma_{i,j} = \frac{1}{m}, \, \gamma_{i,j} \ge 0,
\end{align}
where $\mu(\cdot, \cdot)>0$ is a cost function between two elements, and $m$ and $n$ are the number of elements in each layout.
By solving this problem, we establish a \emph{soft} alignment between the elements denoted as $\gamma^\ast$. 

We here define the cost function based on similarities of bounding boxes' position and labels as follows:
\begin{equation}
    \mu(e_i, \hat{e}_j) = 1 - \frac{\delta_{\mathrm{bbox}}(b_i, \hat{b}_j) + \delta_{\mathrm{label}}(c_i, \hat{c}_j)}{2}.
\end{equation}
The positional similarity is computed as
\begin{equation}
\delta_{\mathrm{bbox}}(b_i, \hat{b}_j) = \frac{1+\mathrm{GIoU}(b_i, \hat{b}_j)}{2},
\end{equation}
where $\mathrm{GIoU}(\cdot, \cdot) \in [-1, 1]$ is the generalized IoU~\cite{Rezatofighi_2018_CVPR}, thus this function ranges from zero to one.
The value gets closer to one when two bounding boxes overlap more.
The label similarity is an indicator function
\begin{equation}
\delta_{\mathrm{label}}(c_i, \hat{c}_j) = 
\begin{cases}
1 & (c_i = \hat{c}_j) \\
0 & (c_i \neq \hat{c}_j)
\end{cases}.
\end{equation}
Unlike prior measures, elements in similar positions still be able to get rewards, even if they are in different categories.

After obtaining the soft alignment of elements, we total the transportation cost as a dissimilarity between $\mathcal{L}$ and $\hat{\mathcal{L}}$:
\begin{equation}
\mathrm{EMD}(\mathcal{L}, \hat{\mathcal{L}}) = \sum_{i,j} \gamma^\ast_{i,j} \mu(e_i, \hat{e}_j).
\end{equation}
$\mathrm{EMD}(\cdot, \cdot)$ is a dissimilarity that ranges from zero to one.
We transform this dissimilarity into a \ourmeasure{} similarity as:
\begin{equation}
    \mathrm{LTSim}(\mathcal{L}, \hat{\mathcal{L}}; \sigma) = \exp\left( - \frac{\mathrm{EMD}(\mathcal{L}, \hat{\mathcal{L}})}{\sigma}\right),
\end{equation}
where $\sigma>0$ is a scaling parameter that we introduce for the collection-level measure described in the next section. For layout-level similarity, we simply set $\sigma=1.0$.

Unlike traditional measures such as DocSim, \maxiou{}, and DocEMD, which focus on element similarity within the same category, our measure employs a soft matching strategy, \ie, many-to-many cross-category matching.
This enables our measure to quantify the differences in layouts with varying numbers of elements or category mismatches.

\subsection{Collection-level Similarity}
Tasks like unconditional layout generation use collection-level evaluation such as \maxiou{} and FID.
We extend \ourmeasure{} for collection-level comparison using Maximum Mean Discrepancy (MMD)~\cite{MMD_gretton12a}. 
MMD is a common choice for distribution comparison~\cite{bińkowski2018demystifying}.
Given a pairwise affinity function, MMD assesses the discrepancy of collections by comparing the average similarity within and across collections.

Here, we consider a collection of real layouts $\mathcal{C} = \{\mathcal{L}_1, \ldots, \mathcal{L}_s\}$, and a generated one $\hat{\mathcal{C}} = \{\hat{\mathcal{L}}_1, \ldots, \hat{\mathcal{L}}_t\}$.
With an unbiased empirical estimator of the squared MMD, the discrepancy between $\mathcal{C}$ and $\mathcal{C}'$ is evaluated as:
\begin{equation}
\begin{aligned}
    \widehat{\mathrm{MMD}^2}(\mathcal{C}, \hat{\mathcal{C}}) = & \frac{1}{s(s-1)}\sum_{i \neq j}^s k(\mathcal{L}_i, \mathcal{L}_j) \\
    & + \frac{1}{t(t-1)}\sum_{i \neq j}^t k(\hat{\mathcal{L}}_i, \hat{\mathcal{L}}_j) \\
    & - \frac{2}{st} \sum_{i=1}^{s} \sum_{j=1}^{t} k(\mathcal{L}_i, \hat{\mathcal{L}}_j).
\end{aligned}
\end{equation}
We define an affinity function as:
\begin{equation}
    k(\mathcal{L}_i, \mathcal{L}_j) = \mathrm{LTSim}(\mathcal{L}_i, \mathcal{L}_j; \sigma),
\end{equation}
where we set the scaling parameter $\sigma$ to a median of $\mathrm{EMD}(\cdot, \cdot)$ on real layout pairs~\cite{sriperumbudur2009kernel}.
We have to assess every layout pair within and across collections, but parallel computing lets us efficiently get an MMD measurement\footnote{\ourmeasure{} can process \textasciitilde460 pairs/sec.\ on a laptop with Intel Core i5. We use Google Cloud Dataflow to accelerate the computation of MMD. Details are described in the supplementary material}.

\begin{figure*}[t!]
\centering
    \begin{minipage}[t]{0.48\linewidth}
    \centering
    \textsc{RICO}\\
        \includegraphics[clip,width=\linewidth]{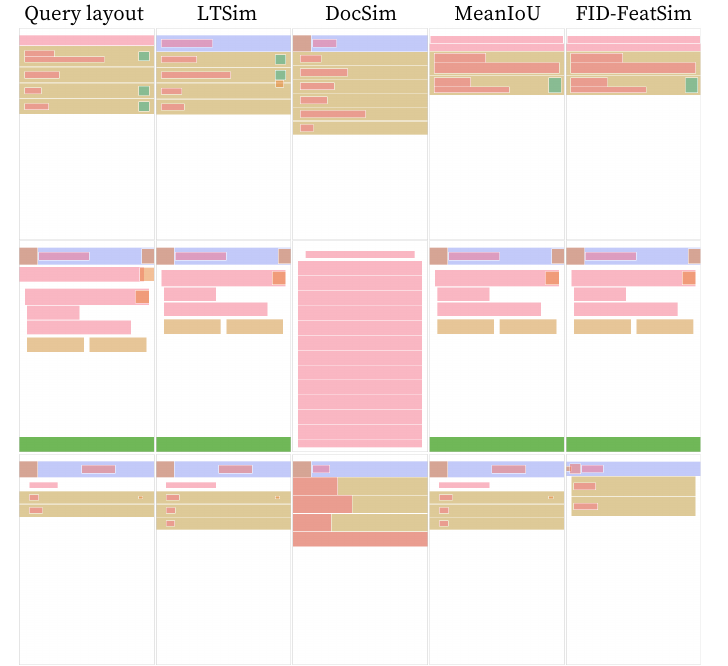}
    \end{minipage}
    \begin{minipage}[t]{0.48\linewidth}
    \centering
    \textsc{PubLayNet}\\
        \includegraphics[clip,width=\linewidth]{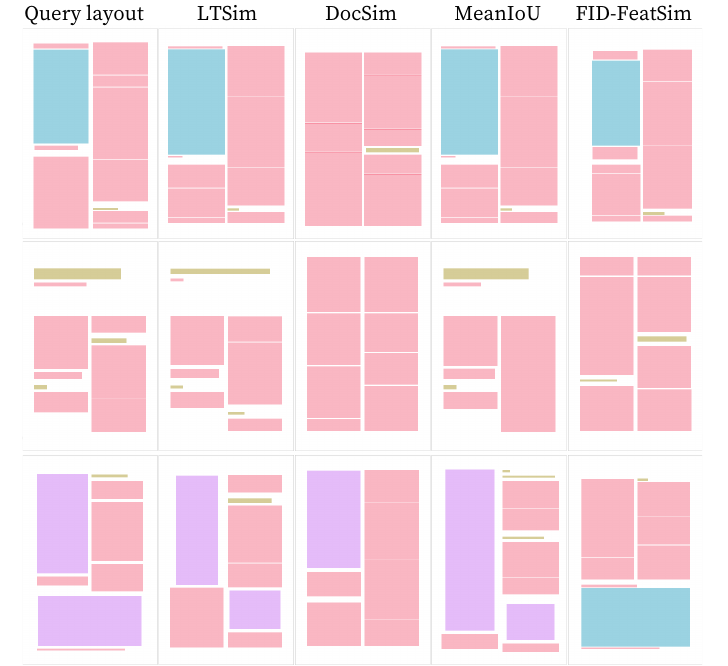}
    \end{minipage}
    \caption{Retrieval examples on \textsc{RICO} and \textsc{PubLayNet}. The leftmost is the query layout and the others are top-1 retrieval by \ourmeasure{}, DocSim, MeanIoU, and \deepsim{}.}
    \label{fig:retrieval}
\end{figure*}
\section{Experiments}
While assessing alignment with human perception may seem like a straightforward method for examining evaluation measures, its reliability for layout similarity is questioned as discussed in \cite{Patil_2021_CVPR}, making subjective evaluation a weak validation tool.
We therefore conducted a series of experiments to investigate key properties of evaluation measures, such as responses to perturbations and collection size effects.
Furthermore, in the supplementary material, we offer a collection of qualitative examples and a thorough comparison of the measures' preferences across diverse layout differences, enabling readers to verify the behavior of these measures.

\paragraph{Dataset}
We use \textsc{RICO}~\cite{rico} and \textsc{PubLayNet}~\cite{publaynet}.
Experiments are done with their validation sets except for the one in \cref{sec:eval_gen_models} where we re-evaluate generation models on the test sets.
\textsc{RICO} provides layouts of mobile apps' user interfaces.
Each element is labeled as one of 25 element categories such as text button, image, and icon. 
Its validation and test sets include 2,109 and 4,218 layouts, respectively. 
\textsc{PubLayNet} is a dataset of scientific papers. 
Each element is labeled with one of five element categories, such as table, image, and text. 
Its validation and test sets include 16,619 and 11,142 layouts, respectively.
The data splits above may differ from the official ones due to the data filtering, \eg, excluding samples with more than 25 elements, as in~\cite{inoue2023layout}.

\paragraph{Compared measures}
We compare \ourmeasure{}/-MMD with DocSim, $\mathrm{MaxIoU}_{\mathrm{\beta}}$, \maxiou{}, MeanIoU, and FID. We test FID in~\cite{inoue2023layout} based on learned features that encode boxes' spatial representation and labels.
We also investigate the capability of the learned feature space used in FID~\cite{inoue2023layout}. 
Specifically, we compute the similarity between two layout features as $\exp(-\|\mathbf{x} - \hat{\mathbf{x}}\|^2_2/2\sigma^2)$ where $\mathbf{x}, \hat{\mathbf{x}} \in \mathbb{R}^{d}$ are leaned features and $\sigma$ is the median of the squared euclidean distance of every layout pairs in the dataset.
We call this measure \deepsim{}.
FID and \deepsim{} serve as representatives of data-driven evaluations, whereas DocSim, $\mathrm{MaxIoU}_{\mathrm{\beta}}$, \maxiou{}, and MeanIoU stand as examples of manually designed methods.

\subsection{Layout-level Similarity}
\subsubsection{Layout Retrieval}
To qualitatively demonstrate how each layout similarity measure behaves, we show layout retrieval results in \cref{fig:retrieval}.
For each query layout, we retrieve the top nearest sample.
The result by $\mathrm{MaxIoU}_{\mathrm{\beta}}$ is not displayed as most queries do not have comparable samples due to its requirement for category multiset matching.
Notably, DocSim's results significantly differ from the others, while the other measures yield similar results with minor differences.
We assume that this difference stems from DocSim's emphasis on element sizes.
We observe that \deepsim{} sometimes fails to retrieve similar layouts on \textsc{PubLayNet}.
A possible explanation for these failures is that the learned features used in \deepsim{} lose their representation capability around the space where training samples are not enough, which leads to unstable performance.
This is a primary challenge of the representation learning-based approach.
The retrieval results on \textsc{RICO} also demonstrate how each measure considers category matching.
MeanIoU retrieves layouts with identical element categories in different placements, while ours retrieves layouts with similarly positioned elements and some label mismatches.
This is because ours does not view label mismatch as a decisive factor, instead, it pays attention to both positional and label similarity.

\subsubsection{Correlation between Measures}
\label{sec:corr}
We do not intend to introduce a drastic change in layout evaluation criteria, but rather to broaden the scope of assessable pairs.
Therefore, we verify the agreement between ours and the existing measures.
In this experiment, we rank layout pairs using the similarity measures, and then calculate the rank correlation between each pair of rankings.
We ensure that each layout pair has the same category multiset so that $\mathrm{MaxIoU}_{\mathrm{\beta}}$ can compute its similarity.
As a rank correlation measure, we use Kendall's $\tau$.
Strong agreement leads to values near 1, while disagreement leads to -1.
The result shown in \cref{tab:corr} suggests that, for layout pairs whose element labels are identical, \ourmeasure{} positively agrees with existing measures, except for DocSim.
The low correlation between DocSim and other metrics implies that DocSim may lead to inconsistent conclusions with other measures.
On \textsc{PubLayNet}, while the measures still show positive correlations, the values drop overall.

\begin{table}[t!]
    \centering
\begin{tabular}{@{}rccccc@{}}
 & \rotatebox{70}{\ourmeasure{}} & \rotatebox{70}{$\mathrm{Max.IoU}_{\beta}$} & \rotatebox{70}{MeanIoU} & \rotatebox{70}{\shortstack{FID-\\FeatSim}} & \rotatebox{70}{DocSim} \\
\midrule
\multicolumn{1}{l}{\small\color{gray}\textsc{RICO}} & & & & & \\
\ourmeasure{} & {\cellcolor[HTML]{000000}} \color[HTML]{F1F1F1} {\cellcolor{white}} \color{white} --- & {\cellcolor[HTML]{08306B}} \color[HTML]{F1F1F1} 0.99 & {\cellcolor[HTML]{083B7C}} \color[HTML]{F1F1F1} 0.97 & {\cellcolor[HTML]{4B98CA}} \color[HTML]{F1F1F1} 0.80 & {\cellcolor[HTML]{B9D6EA}} \color[HTML]{000000} 0.65 \\
\maxiou{}$_{\mathrm{\beta}}$ & {\cellcolor[HTML]{08306B}} \color[HTML]{F1F1F1} 0.99 & {\cellcolor[HTML]{000000}} \color[HTML]{F1F1F1} {\cellcolor{white}} \color{white} --- & {\cellcolor[HTML]{083979}} \color[HTML]{F1F1F1} 0.97 & {\cellcolor[HTML]{4E9ACB}} \color[HTML]{F1F1F1} 0.79 & {\cellcolor[HTML]{B8D5EA}} \color[HTML]{000000} 0.65 \\
MeanIoU & {\cellcolor[HTML]{083B7C}} \color[HTML]{F1F1F1} 0.97 & {\cellcolor[HTML]{083979}} \color[HTML]{F1F1F1} 0.97 & {\cellcolor[HTML]{000000}} \color[HTML]{F1F1F1} {\cellcolor{white}} \color{white} --- & {\cellcolor[HTML]{58A1CF}} \color[HTML]{F1F1F1} 0.78 & {\cellcolor[HTML]{BFD8ED}} \color[HTML]{000000} 0.64 \\
\deepsim{} & {\cellcolor[HTML]{4B98CA}} \color[HTML]{F1F1F1} 0.80 & {\cellcolor[HTML]{4E9ACB}} \color[HTML]{F1F1F1} 0.79 & {\cellcolor[HTML]{58A1CF}} \color[HTML]{F1F1F1} 0.78 & {\cellcolor[HTML]{000000}} \color[HTML]{F1F1F1} {\cellcolor{white}} \color{white} --- & {\cellcolor[HTML]{F7FBFF}} \color[HTML]{000000} 0.52 \\
DocSim & {\cellcolor[HTML]{B9D6EA}} \color[HTML]{000000} 0.65 & {\cellcolor[HTML]{B8D5EA}} \color[HTML]{000000} 0.65 & {\cellcolor[HTML]{BFD8ED}} \color[HTML]{000000} 0.64 & {\cellcolor[HTML]{F7FBFF}} \color[HTML]{000000} 0.52 & {\cellcolor[HTML]{000000}} \color[HTML]{F1F1F1} {\cellcolor{white}} \color{white} --- \\

\midrule
\multicolumn{1}{l}{\small\color{gray}\textsc{PubLayNet}} & & & & & \\
\ourmeasure{} & {\cellcolor[HTML]{000000}} \color[HTML]{F1F1F1} {\cellcolor{white}} \color{white} --- & {\cellcolor[HTML]{08306B}} \color[HTML]{F1F1F1} 0.77 & {\cellcolor[HTML]{66ABD4}} \color[HTML]{F1F1F1} 0.55 & {\cellcolor[HTML]{C2D9EE}} \color[HTML]{000000} 0.44 & {\cellcolor[HTML]{56A0CE}} \color[HTML]{F1F1F1} 0.58 \\
\maxiou{}$_{\mathrm{\beta}}$ & {\cellcolor[HTML]{08306B}} \color[HTML]{F1F1F1} 0.77 & {\cellcolor[HTML]{000000}} \color[HTML]{F1F1F1} {\cellcolor{white}} \color{white} --- & {\cellcolor[HTML]{539ECD}} \color[HTML]{F1F1F1} 0.58 & {\cellcolor[HTML]{CBDEF1}} \color[HTML]{000000} 0.42 & {\cellcolor[HTML]{4594C7}} \color[HTML]{F1F1F1} 0.60 \\
MeanIoU & {\cellcolor[HTML]{66ABD4}} \color[HTML]{F1F1F1} 0.55 & {\cellcolor[HTML]{539ECD}} \color[HTML]{F1F1F1} 0.58 & {\cellcolor[HTML]{000000}} \color[HTML]{F1F1F1} {\cellcolor{white}} \color{white} --- & {\cellcolor[HTML]{F7FBFF}} \color[HTML]{000000} 0.32 & {\cellcolor[HTML]{D7E6F5}} \color[HTML]{000000} 0.39 \\
\deepsim{} & {\cellcolor[HTML]{C2D9EE}} \color[HTML]{000000} 0.44 & {\cellcolor[HTML]{CBDEF1}} \color[HTML]{000000} 0.42 & {\cellcolor[HTML]{F7FBFF}} \color[HTML]{000000} 0.32 & {\cellcolor[HTML]{000000}} \color[HTML]{F1F1F1} {\cellcolor{white}} \color{white} --- & {\cellcolor[HTML]{AFD1E7}} \color[HTML]{000000} 0.47 \\
DocSim & {\cellcolor[HTML]{56A0CE}} \color[HTML]{F1F1F1} 0.58 & {\cellcolor[HTML]{4594C7}} \color[HTML]{F1F1F1} 0.60 & {\cellcolor[HTML]{D7E6F5}} \color[HTML]{000000} 0.39 & {\cellcolor[HTML]{AFD1E7}} \color[HTML]{000000} 0.47 & {\cellcolor[HTML]{000000}} \color[HTML]{F1F1F1} {\cellcolor{white}} \color{white} --- \\
\bottomrule

\end{tabular}
    \caption{Rank correlation between the measures. We rank layout pairs by their similarity using each measure and compute Kendall's $\tau$ across them. The darker colors depict a strong correlation.}
    \label{tab:corr}
\end{table}

\subsection{Collection-level Similarity}
\subsubsection{Reliability Analysis}
\label{sec:response}
For a reliable comparison of layout generation models, measures should accurately gauge the degree of difference between collections. 
To validate this, we perturb a layout dataset to create layout collections with varying degrees of differences from the original and examine whether the measures accurately reflect the degree of difference.
We use two types of perturbations: i) \textit{Positional noise} shifts the bounding box position within a normalized offset uniformly sampled from a range from 0 to 0.1, and 
ii) \textit{Label noise} replaces a label with a different one.
These noises are independently applied to elements at a certain rate. 
We then calculate the similarity or discrepancy between the original and the perturbed datasets.
We adjust the perturbation injection rate from 0.1 to 0.5 and run 10 trials for each setting.
Examples of perturbed layouts are in the supplementary material.

\Cref{fig:response_to_noise} shows the result on \textsc{RICO} validation set that has 2,109 layouts.
Each box represents the range of measures with 10 trials.
We observe \maxiou{} values overlap across different perturbation injection rates. 
This implies that \maxiou{} can potentially yield unreliable results, as it might assign higher similarity to a pair of layout collections with larger differences than the one with smaller differences.
FID shows stability but cannot differentiate collections with a perturbation rate of 10\% from those with 20\%.
On the other hand, \ourmeasure{}-MMD successfully distinguishes collections with different degrees of difference.

\begin{figure*}
    \begin{minipage}[t]{0.65\linewidth}
        \centering
        \includegraphics[clip,width=\linewidth]{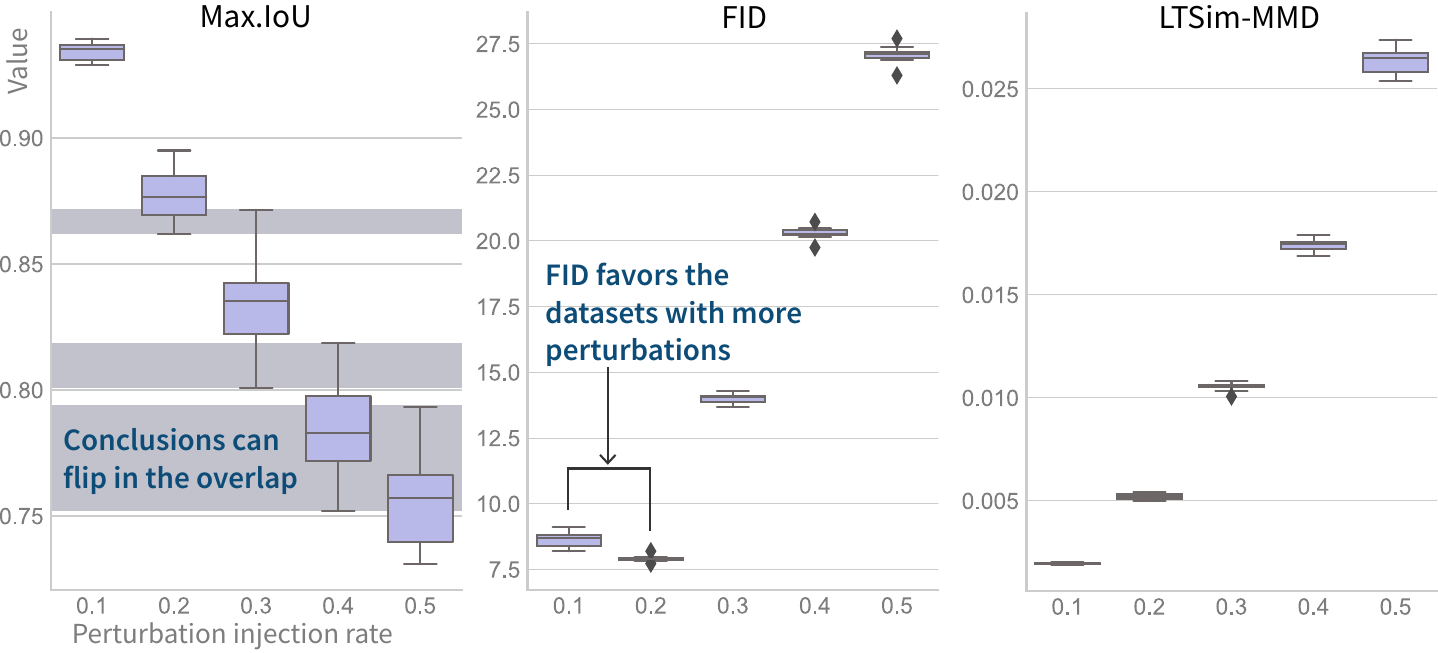}
        \caption{These box plots show the responses to perturbations by \maxiou{}, FID, and \ourmeasure{}-MMD. The overlaps imply that the measure may fail to distinguish the quality of layout collections.}
        \label{fig:response_to_noise}
    \end{minipage}
    \hspace{0.03\linewidth}
    \begin{minipage}[t]{0.3\linewidth}
        \centering
        \includegraphics[clip,width=\linewidth]{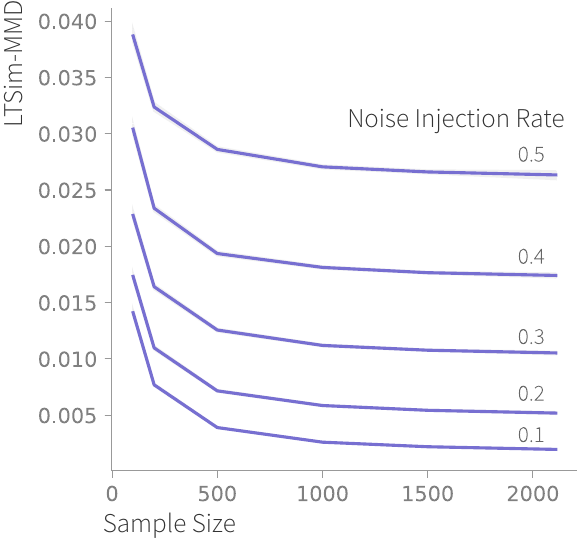}
        \caption{\ourmeasure{}-MMD on varying collection size. This measure still provides a good estimate even on the halved dataset.}
        \label{fig:sample_size}
    \end{minipage}
\end{figure*}

\subsubsection{Effect of Collection Size}
Parallel computing can accelerate MMD calculations, but it remains relatively costly. 
Using a smaller sample size for a preliminary estimate will facilitate quick model development. 
We examine how the size of the layout collection impacts \ourmeasure{}-MMD estimation. 
We sample a subset of the perturbed datasets 10 times for each and reassess \ourmeasure{}-MMD.
The size of the reference collection remains unchanged.
\Cref{fig:sample_size} shows that \ourmeasure{}-MMD consistently ranks perturbed collections, with only minor changes even when the dataset size is halved with a 42\% decrease in computation costs in this setting.
This result suggests that our measure works even with small datasets, making it useful for model validation during training.

\begin{table*}[t!]
    \centering
\begin{tabular}{lrccc|cccc}
\toprule
 &  & \multicolumn{3}{c|}{Unconditional Gen.} & \multicolumn{4}{c}{Label-conditioned Gen.} \\
 &  & \maxiou{}$\uparrow$ & FID$\downarrow$ & \shortstack{\ourmeasure{}\\-MMD}$\downarrow$ & DocSim$\uparrow$ & MeanIoU$\uparrow$ & \maxiou{}$\uparrow$ & \ourmeasure{}$\uparrow$ \\
\midrule
\multirow[c]{5}{*}{\textsc{RICO}} & BART~\cite{lewis-etal-2020-bart} & {\cellcolor[HTML]{D6E6F4}} \color[HTML]{000000} 38.06 & {\cellcolor[HTML]{ABD0E6}} \color[HTML]{000000} 11.90 & {\cellcolor[HTML]{ABD0E6}} \color[HTML]{000000} 1.08 & {\cellcolor[HTML]{D6E6F4}} \color[HTML]{000000} 16.25 & {\cellcolor[HTML]{D6E6F4}} \color[HTML]{000000} 29.01 & {\cellcolor[HTML]{D6E6F4}} \color[HTML]{000000} 24.93 & {\cellcolor[HTML]{D6E6F4}} \color[HTML]{000000} 75.46 \\
 & BLT~\cite{kong_blt_2022} & {\cellcolor[HTML]{F7FBFF}} \color[HTML]{000000} 21.98 & {\cellcolor[HTML]{F7FBFF}} \color[HTML]{000000} 178.19 & {\cellcolor[HTML]{F7FBFF}} \color[HTML]{000000} 26.21 & {\cellcolor[HTML]{F7FBFF}} \color[HTML]{000000} 13.34 & {\cellcolor[HTML]{F7FBFF}} \color[HTML]{000000} 22.93 & {\cellcolor[HTML]{F7FBFF}} \color[HTML]{000000} 20.46 & {\cellcolor[HTML]{F7FBFF}} \color[HTML]{000000} 73.64 \\
 & LayoutDM~\cite{inoue2023layout} & {\cellcolor[HTML]{6AAED6}} \color[HTML]{F1F1F1} 48.36 & {\cellcolor[HTML]{3787C0}} \color[HTML]{F1F1F1} 6.65 & {\cellcolor[HTML]{3787C0}} \color[HTML]{F1F1F1} 0.26 & {\cellcolor[HTML]{6AAED6}} \color[HTML]{F1F1F1} 16.79 & {\cellcolor[HTML]{3787C0}} \color[HTML]{F1F1F1} 32.41 & {\cellcolor[HTML]{3787C0}} \color[HTML]{F1F1F1} 27.69 & {\cellcolor[HTML]{3787C0}} \color[HTML]{F1F1F1} 76.24 \\
 & MaskGIT~\cite{Chang_2022_CVPR} & {\cellcolor[HTML]{3787C0}} \color[HTML]{F1F1F1} 57.69 & {\cellcolor[HTML]{D6E6F4}} \color[HTML]{000000} 52.11 & {\cellcolor[HTML]{D6E6F4}} \color[HTML]{000000} 3.93 & {\cellcolor[HTML]{3787C0}} \color[HTML]{F1F1F1} 16.87 & {\cellcolor[HTML]{6AAED6}} \color[HTML]{F1F1F1} 30.41 & {\cellcolor[HTML]{6AAED6}} \color[HTML]{F1F1F1} 26.23 & {\cellcolor[HTML]{6AAED6}} \color[HTML]{F1F1F1} 75.92 \\
 & VQDiffusion~\cite{Gu_2022_CVPR} & {\cellcolor[HTML]{ABD0E6}} \color[HTML]{000000} 43.42 & {\cellcolor[HTML]{6AAED6}} \color[HTML]{F1F1F1} 7.46 & {\cellcolor[HTML]{6AAED6}} \color[HTML]{F1F1F1} 0.58 & {\cellcolor[HTML]{ABD0E6}} \color[HTML]{000000} 16.46 & {\cellcolor[HTML]{ABD0E6}} \color[HTML]{000000} 29.01 & {\cellcolor[HTML]{ABD0E6}} \color[HTML]{000000} 25.20 & {\cellcolor[HTML]{ABD0E6}} \color[HTML]{000000} 75.57 \\
\midrule
\multirow[c]{5}{*}{\textsc{PubLayNet}} & BART~\cite{lewis-etal-2020-bart} & {\cellcolor[HTML]{6AAED6}} \color[HTML]{F1F1F1} 45.09 & {\cellcolor[HTML]{ABD0E6}} \color[HTML]{000000} 16.60 & {\cellcolor[HTML]{ABD0E6}} \color[HTML]{000000} 1.89 & {\cellcolor[HTML]{6AAED6}} \color[HTML]{F1F1F1} 13.08 & {\cellcolor[HTML]{6AAED6}} \color[HTML]{F1F1F1} 27.25 & {\cellcolor[HTML]{6AAED6}} \color[HTML]{F1F1F1} 32.00 & {\cellcolor[HTML]{6AAED6}} \color[HTML]{F1F1F1} 76.27 \\
 & BLT~\cite{kong_blt_2022} & {\cellcolor[HTML]{F7FBFF}} \color[HTML]{000000} 32.07 & {\cellcolor[HTML]{F7FBFF}} \color[HTML]{000000} 116.81 & {\cellcolor[HTML]{F7FBFF}} \color[HTML]{000000} 26.59 & {\cellcolor[HTML]{F7FBFF}} \color[HTML]{000000} 10.47 & {\cellcolor[HTML]{F7FBFF}} \color[HTML]{000000} 24.53 & {\cellcolor[HTML]{F7FBFF}} \color[HTML]{000000} 21.49 & {\cellcolor[HTML]{F7FBFF}} \color[HTML]{000000} 74.77 \\
 & LayoutDM~\cite{inoue2023layout} & {\cellcolor[HTML]{ABD0E6}} \color[HTML]{000000} 39.05 & {\cellcolor[HTML]{3787C0}} \color[HTML]{F1F1F1} 13.91 & {\cellcolor[HTML]{3787C0}} \color[HTML]{F1F1F1} 0.85 & {\cellcolor[HTML]{D6E6F4}} \color[HTML]{000000} 12.69 & {\cellcolor[HTML]{ABD0E6}} \color[HTML]{000000} 26.38 & {\cellcolor[HTML]{D6E6F4}} \color[HTML]{000000} 31.04 & {\cellcolor[HTML]{D6E6F4}} \color[HTML]{000000} 75.74 \\
 & MaskGIT~\cite{Chang_2022_CVPR} & {\cellcolor[HTML]{3787C0}} \color[HTML]{F1F1F1} 51.40 & {\cellcolor[HTML]{D6E6F4}} \color[HTML]{000000} 32.27 & {\cellcolor[HTML]{D6E6F4}} \color[HTML]{000000} 7.73 & {\cellcolor[HTML]{3787C0}} \color[HTML]{F1F1F1} 13.22 & {\cellcolor[HTML]{3787C0}} \color[HTML]{F1F1F1} 27.87 & {\cellcolor[HTML]{ABD0E6}} \color[HTML]{000000} 31.92 & {\cellcolor[HTML]{3787C0}} \color[HTML]{F1F1F1} 76.58 \\
 & VQDiffusion~\cite{Gu_2022_CVPR} & {\cellcolor[HTML]{D6E6F4}} \color[HTML]{000000} 37.76 & {\cellcolor[HTML]{6AAED6}} \color[HTML]{F1F1F1} 15.41 & {\cellcolor[HTML]{6AAED6}} \color[HTML]{F1F1F1} 0.93 & {\cellcolor[HTML]{ABD0E6}} \color[HTML]{000000} 12.79 & {\cellcolor[HTML]{D6E6F4}} \color[HTML]{000000} 26.34 & {\cellcolor[HTML]{3787C0}} \color[HTML]{F1F1F1} 32.09 & {\cellcolor[HTML]{ABD0E6}} \color[HTML]{000000} 75.89 \\
\bottomrule
\end{tabular}

    \caption{Evaluation results on unconditional and label-conditioned generation. The darker colors indicate the higher ranks. The measurement values except FID are scaled by 100 for visibility.}
    \label{tab:eval_generated}
\end{table*}
\subsection{Evaluating Generated Layouts}
\label{sec:eval_gen_models}
We evaluate generated layouts on tasks in~\cite{inoue2023layout}.
We use the test sets in this experiment.

\paragraph{Unconditional generation} Models generate layouts so that the generated layouts look like real ones. 
We test collection-level measures, \ie, \maxiou{}, FID, \ourmeasure{}-MMD, on this task.
\Cref{tab:eval_generated} shows the result. 
FID and \ourmeasure{}-MMD both rank the same model as the best, while \maxiou{} prefers a different one. 
\maxiou{}, designed for label-conditioned generation, evaluates only a small subset of layouts whose label multiset is identical to one of the reference samples. 
In \textsc{RICO} experiments, \maxiou{} evaluates only 18\% of generated layouts on average among all results. 
This skews the layout distribution under evaluation, leading to a different conclusion compared to other collection-level measures.
This result shows the risk of using \maxiou{} for tasks other than label-conditioned generation.
We observe a similar trend on \textsc{PubLayNet}.

\paragraph{Label-conditioned generation} 
In this task, given element categories, \eg, one image and two text buttons, the model predicts the sizes and positions for them. 
We test layout-level measures and \maxiou{}, with results displayed in \cref{tab:eval_generated}.
MeanIoU, \maxiou{} and \ourmeasure{} yield identical model rankings on \textsc{RICO}.
In this task, all samples are properly involved in \maxiou{}, avoiding the issue seen in unconditional generation.
As discussed, DocSim places significant emphasis on element sizes, leading to a ranking different from other measures.
We observe minor disagreement across the measures on \textsc{PubLayNet} as expected from the lower correlations in \cref{tab:corr}.

\section{Discussions}
\paragraph{Which measures should we use?}
\textbf{DocSim}: Its emphasis on element sizes can result in unintuitive similarities. There's no strong research basis for using it.
\textbf{\maxiou{}}: It is suitable for label-conditioned generation, but not recommended for other tasks involving label generation such as unconditional generation.
\textbf{FID}: FID often yields similar results to ours. However, its effectiveness relies on a feature extractor, which shows questionable results in \cref{sec:response}. Ensuring the feature extractor's reliability is crucial, yet highly challenging.
\textbf{\ourmeasure{}/-MMD}: We suggest using them as a primary measure. As \ourmeasure{} is designed to handle layouts with different label sets, it covers tasks involving label prediction.
For collection-level comparison, \ourmeasure{}-MMD is beneficial. 
When its computational cost is not affordable, we suggest a preliminary evaluation using a small subset and a full dataset evaluation for the final assessment.

\paragraph{Limitations}
The main challenge is \ourmeasure{}-MMD's computational cost.
We need to evaluate every pair within and across the collections.
Although parallel computing alleviates the issue, reducing the cost will be an important topic when precise evaluation with a massive dataset is required.

Layout generation aims to create plausible and novel layouts. 
However, current evaluation measures, including ours, may favor layouts that are simply duplicates of training samples. 
It would be interesting to develop an evaluation measure that penalizes the generation of layouts overly similar to reference ones.

\section{Conclusion}
The maintenance of evaluation tools is crucial for ensuring the sound development of the field. 
We review the layout generation evaluation's similarity measures and uncover their challenges. 
Based on the findings, we introduce a new measure, \ourmeasure{}. 
Our measure quantifies various layout differences often missed by existing measures. 
We also extend this measure for collection-level comparisons. 
Empirical results show that our measure offers a more reliable comparison tool suitable for a range of layout generation tasks.

{
    \small
    \bibliographystyle{ieeenat_fullname}
    \bibliography{main}
}

\end{document}


\maketitle
\section{Evaluations in Recent Papers}
\Cref{tab:eval_measures_used} is a list of recent layout generation papers and the evaluation methods used in their experiments.
We observe that the use of FID is quite common.
However, it is important to recognize that the FID measure may be different due to the difference in layout feature extractors.

\newcommand{\na}{\textcolor{lightgray}{N/A}}
\begin{table}[h!]
    \centering
    \begin{tabular}{rL{4cm}L{4cm}} \hline 
         &  Conditional Generation& Unconditional Generation\\ \hline 
         LayoutGAN~\cite{li_layoutgan_2021} & \na & Align., Overlap, \\
         LayoutVAE~\cite{jyothi2019layoutvae} & \na & NLL, IoU variant \\
         NDN~\cite{lee_neural_2020} & FID, Align., Const. consistency & \na \\
         VTN~\cite{Arroyo_2021_CVPR} & \na & Align., Overlap, Wasserstein dist., IoU variant, DocSim \\
         LayoutTransformer~\cite{gupta_layouttransformer_2021} & \na & NLL, Coverage, Overlap \\
         LayoutGAN++~\cite{Kikuchi2021} &  FID, \maxiou{}, Align., Overlap, Const. violation& \na \\ 
         LayoutMCL~\cite{nguyen2021diverse} & \na & FID, Align., fake positive rate \\
         Jiang~\etal~\cite{jiang2022coarse} & \na & Align., Overlap, Wasserstein dist., Chamfer dist. \\
         BLT~\cite{kong_blt_2022} & FID, DocSim, IoU variant, Align., Overlap & IoU variant, Align., Overlap \\
         Yang~\etal~\cite{yang2023layout} & FID, \maxiou{}, Align., Overlap & FID, \maxiou{}, Align., Overlap \\
         PLay~\cite{play_icml23} & FID, G-Usage & \na \\
         LayoutDM~\cite{inoue2023layout} & FID, DocSim, \maxiou{}, Const. violation & FID, Align. \\  
         LayoutDM*~\cite{chai2023layoutdm} & FID, \maxiou{}, IoU variant, Align., Overlap & FID, \maxiou{}, Align., Overlap \\
         LDGM~\cite{hui2023unifying} & FID, \maxiou{}, Align., Overlap & FID, \maxiou{}, Align., Overlap \\
         LayoutFormer++~\cite{jiang_layoutformer_2023} & FID, \maxiou{}, Align., Overlap, Const. violation & FID, \maxiou{}, Align., Overlap \\
         DLT~\cite{levi2023dlt} & FID, DocSim, IoU variant, Align., Overlap & FID, IoU variant, Align., Overlap \\
         LayoutDiffusion~\cite{zhang2023layoutdiffusion} & FID, \maxiou{}, Align., Overlap & FID, \maxiou{}, Align., Overlap \\
         LayoutPrompter~\cite{lin2023layoutprompter} & FID, \maxiou{}, Align., Overlap, Const. violation & FID, \maxiou{}, Align., Overlap \\
         \hline
    \end{tabular}
    \caption{Evaluation measures reported in prior work.}
    \label{tab:eval_measures_used}
\end{table}

\section{Computational Cost of \ourmeasure{}-MMD}
\ourmeasure{} can process \textasciitilde460 pairs/sec.\ on a laptop with Intel Core i5, resulting in \textasciitilde90min.\ to compute MMD on 1K generated and 2K real layouts on RICO. To accelerate this, we use Google Cloud Dataflow. The evaluation of every pair within the RICO validation set, which comprises 4.4 million pairs, is completed in approximately 19 minutes with \$2. We choose the n1-highmem-2 machine type, and set the maximum worker size to 100. Increasing worker size will offer more scalability without a significant cost increase.

\newpage
\section{Perturbed Layout Examples}
\begin{figure}[h!]
    \centering
    \includegraphics[clip,width=\linewidth]{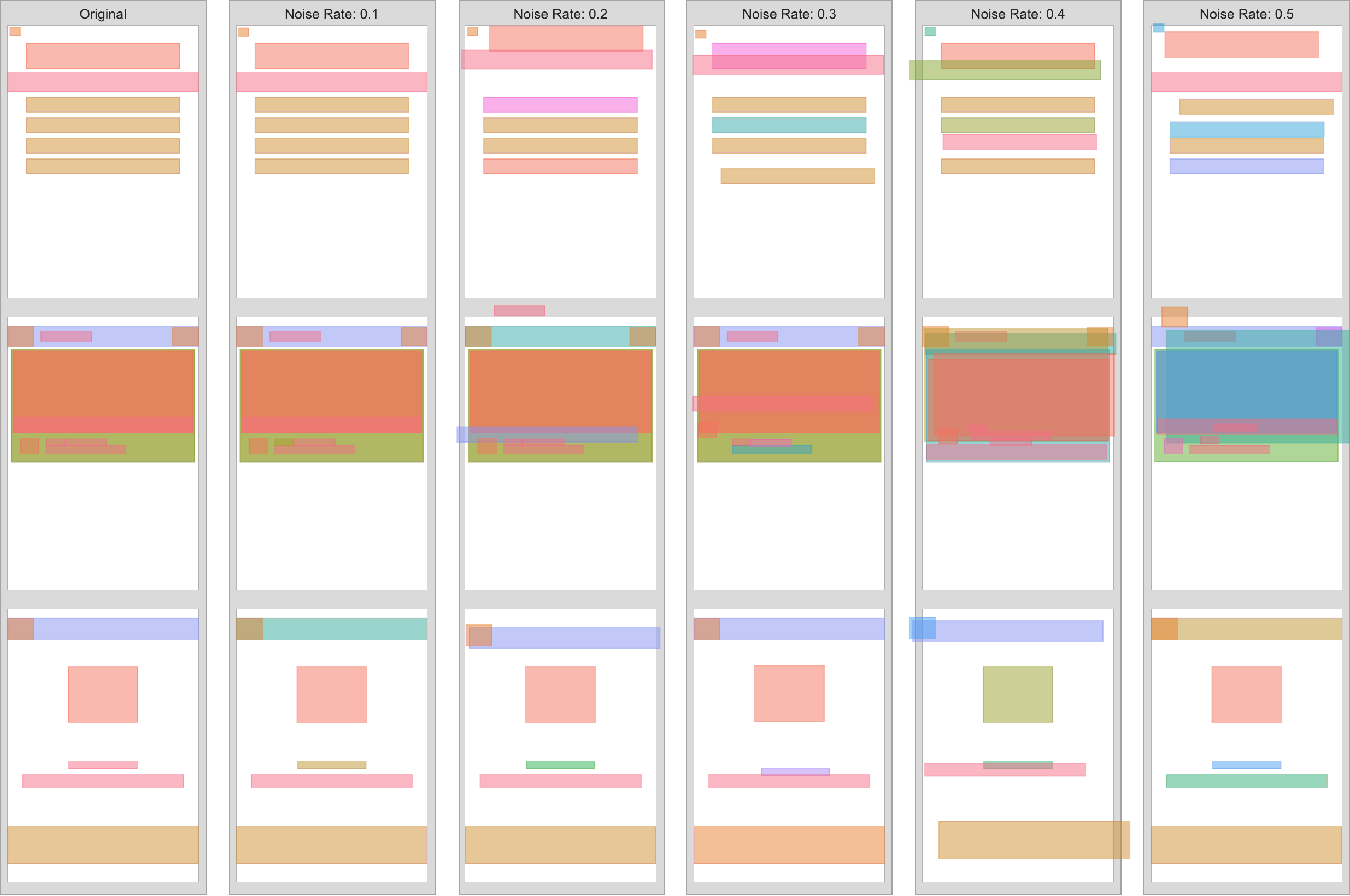}
    \caption{Examples of perturbed layouts used in the reliability analysis experiment \cref{sec:response}. ``Original'' shows an excerpt from the original dataset. With the increase of perturbation injection rate, the dataset changes from left to right.}
    \label{fig:noise_examples}
\end{figure}

\Cref{fig:noise_examples} shows examples of perturbed layout datasets in \cref{sec:response}.
We use two types of perturbations:
\begin{itemize}
    \item \textit{Positional noise} shifts the bounding box position within a normalized offset uniformly sampled from a range from 0 to 0.1, and
    \item \textit{Label noise} replaces a label with a different one.
\end{itemize}
These noises are independently applied to elements at a certain rate. 
We adjust the perturbation injection rate from 0.1 to 0.5.
The leftmost column shows the layouts in the original dataset, while the remaining columns show the perturbed ones.
As the perturbation inject rate gets larger, differences from the original layouts increase.

\newpage
\section{Measures' Preferences}
\begin{figure}
    \begin{minipage}[t]{0.48\textwidth}
    \begin{center}
        \includegraphics[clip,width=0.8\textwidth]{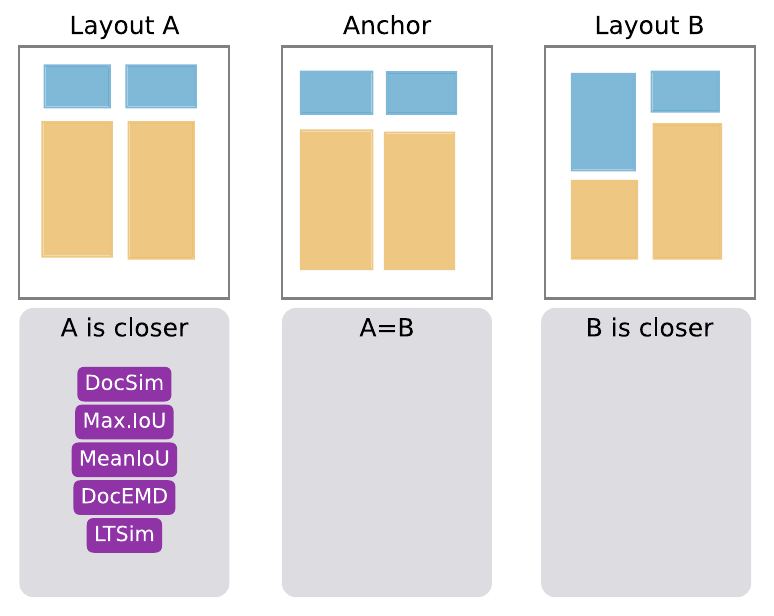}
    \end{center}
    All the measures respond similarly to simple spatial changes in elements.
    \end{minipage}
    \begin{minipage}[t]{0.04\textwidth}
    \vline
    \end{minipage}
    \begin{minipage}[t]{0.48\textwidth}
    \begin{center}
        \includegraphics[clip,width=0.8\textwidth]{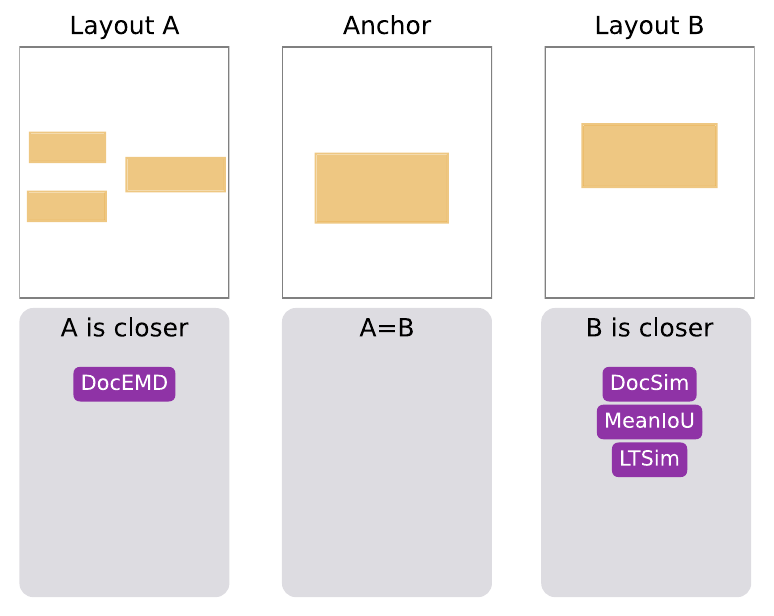}
    \end{center}
    DocEMD views an element as a set of 2D points, thus splitting one element into smaller ones does not significantly reduce the similarity according to DocEMD.
    \end{minipage}
     \begin{minipage}[t]{0.48\textwidth}
        \begin{center}
            \includegraphics[clip,width=0.8\textwidth]{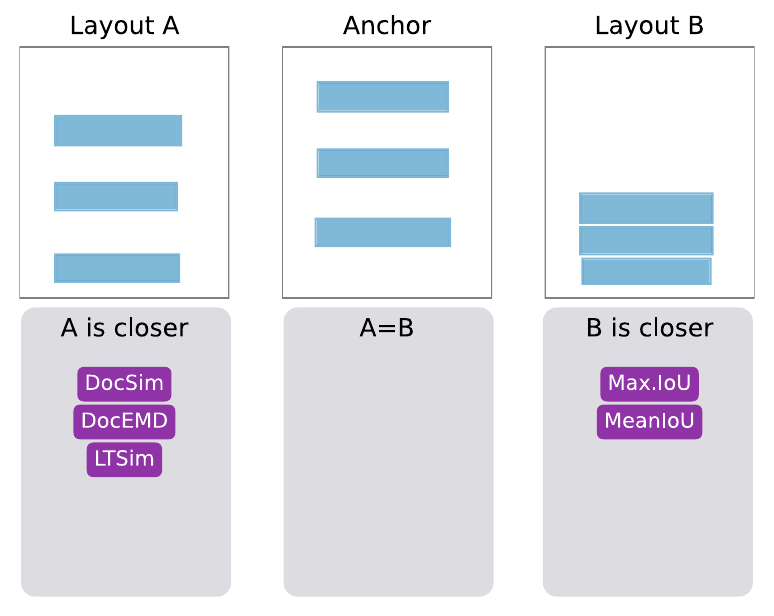}
        \end{center}
        When narrow elements shift slightly, IoU often yields zero, causing a swift decrease in similarity by IoU-based measures like \maxiou{}$_\beta$ and MeanIoU. Our measure uses generalized IoU instead, resulting in a different assessment.
    \end{minipage}
    \begin{minipage}[t]{0.04\textwidth}
    \vline
    \end{minipage}
    \begin{minipage}[t]{0.48\textwidth}
    \begin{center}
        \includegraphics[clip,width=0.8\textwidth]{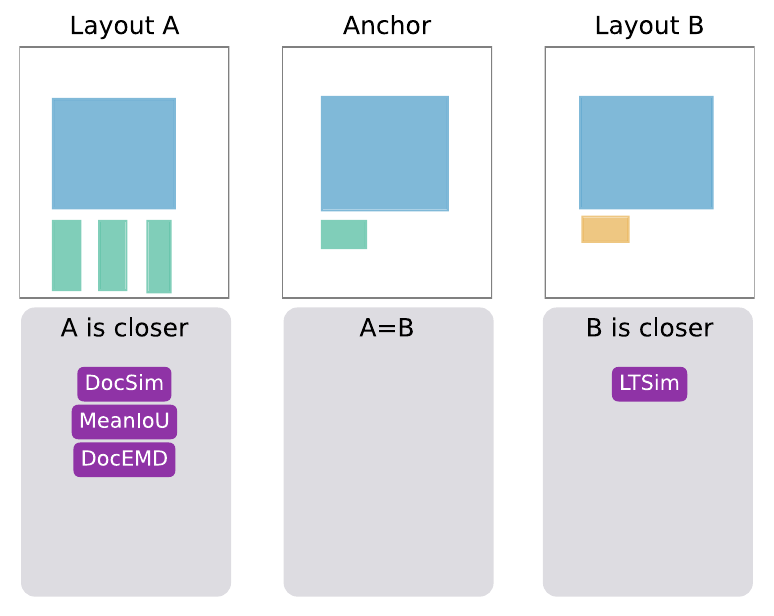}
    \end{center}
    As we discussed in the paper, our measure does not view label mismatch as a decisive factor, thus, prefer one with a highly similar placement in this example.
    \end{minipage}
     \begin{minipage}[t]{0.48\textwidth}
     \begin{center}
         \includegraphics[clip,width=0.8\textwidth]{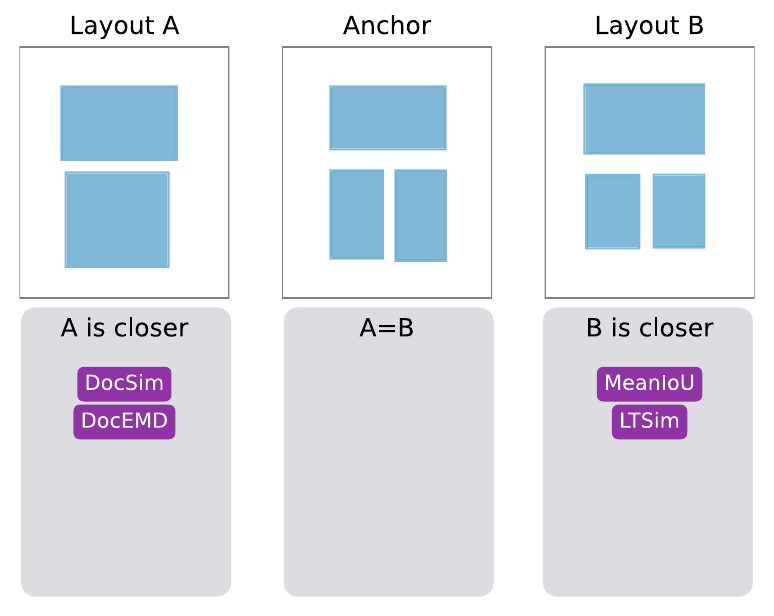}
     \end{center}
     In DocEMD, the similarity is not significantly impacted by merging or splitting elements. In this example, DocEMD looks to favor a layout that has a more common area with the anchor. DocSim often assigns a higher similarity to layouts with larger elements.
    \end{minipage}
    \begin{minipage}[t]{0.04\textwidth}
    \vline
    \end{minipage}
    \begin{minipage}[t]{0.48\textwidth}
    \begin{center}
        \includegraphics[clip,width=0.8\textwidth]{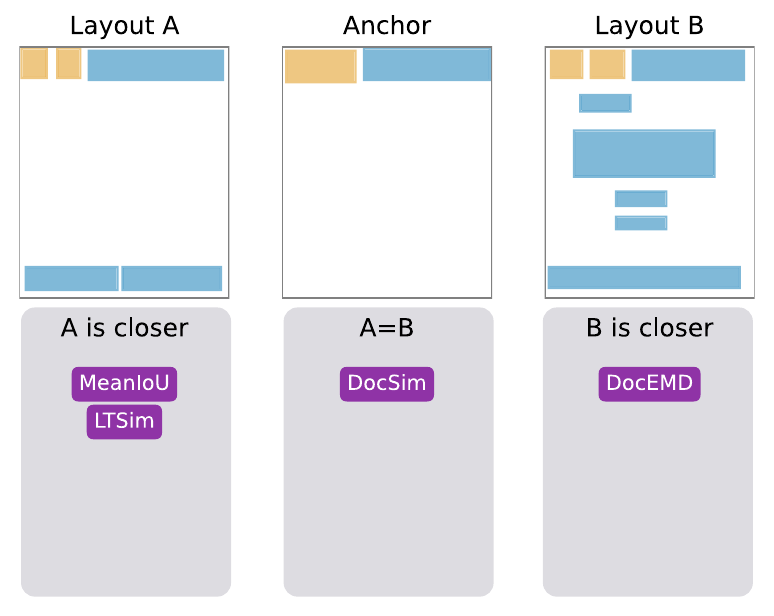}
    \end{center}
    When the difference in element numbers exceeds a threshold, DocSim determines the layouts as entirely dissimilar, resulting in zero similarity.
    \end{minipage}
    \caption{Examples of the measures' preferences.}
    \label{fig:preference}
\end{figure}

\Cref{fig:preference} shows examples of each measure's preferences.
We evaluate whether layout A or B is closer to the anchor using the measures and show each measure's preference. 
Below each example, we provide a brief explanation for these preferences.

\newpage
\section{Evaluation on Layout Completion}
We also report the evaluation results for the layout completion task in \cref{tab:completion}, which requires predicting the remaining elements from a few ones in a reference layout. 
The measures produce slightly different model rankings on \textsc{RICO} and \textsc{PubLayNet}. 
Despite being a conditional generation task, layout completion may have varying numbers and categories of predicted elements compared to reference layouts.
Current measures often struggle with such layout pairs, potentially causing inconsistent results.

\begin{table}[h!]
\centering
\begin{tabular}{lrcccc}
\toprule
 &  & \multicolumn{4}{c}{Layout Completion} \\
 &  & DocSim & MeanIoU & \maxiou{} & \ourmeasure{} \\
\midrule
\multirow[c]{5}{*}{\textsc{RICO}} & BART & {\cellcolor[HTML]{D6E6F4}} \color[HTML]{000000} 7.05 & {\cellcolor[HTML]{ABD0E6}} \color[HTML]{000000} 19.58 & {\cellcolor[HTML]{D6E6F4}} \color[HTML]{000000} 53.02 & {\cellcolor[HTML]{ABD0E6}} \color[HTML]{000000} 62.54 \\
 & BLT & {\cellcolor[HTML]{F7FBFF}} \color[HTML]{000000} 0.76 & {\cellcolor[HTML]{F7FBFF}} \color[HTML]{000000} 12.32 & {\cellcolor[HTML]{F7FBFF}} \color[HTML]{000000} 13.63 & {\cellcolor[HTML]{F7FBFF}} \color[HTML]{000000} 52.73 \\
 & LayoutDM & {\cellcolor[HTML]{6AAED6}} \color[HTML]{F1F1F1} 8.84 & {\cellcolor[HTML]{6AAED6}} \color[HTML]{F1F1F1} 22.08 & {\cellcolor[HTML]{3787C0}} \color[HTML]{F1F1F1} 57.43 & {\cellcolor[HTML]{3787C0}} \color[HTML]{F1F1F1} 63.49 \\
 & MaskGIT & {\cellcolor[HTML]{3787C0}} \color[HTML]{F1F1F1} 11.08 & {\cellcolor[HTML]{3787C0}} \color[HTML]{F1F1F1} 23.08 & {\cellcolor[HTML]{6AAED6}} \color[HTML]{F1F1F1} 55.68 & {\cellcolor[HTML]{D6E6F4}} \color[HTML]{000000} 62.53 \\
 & VQDiffusion & {\cellcolor[HTML]{ABD0E6}} \color[HTML]{000000} 8.25 & {\cellcolor[HTML]{D6E6F4}} \color[HTML]{000000} 19.14 & {\cellcolor[HTML]{ABD0E6}} \color[HTML]{000000} 53.97 & {\cellcolor[HTML]{6AAED6}} \color[HTML]{F1F1F1} 62.72 \\
 \midrule
\multirow[c]{5}{*}{\textsc{PubLayNet}} & BART & {\cellcolor[HTML]{D6E6F4}} \color[HTML]{000000} 7.56 & {\cellcolor[HTML]{3787C0}} \color[HTML]{F1F1F1} 31.01 & {\cellcolor[HTML]{6AAED6}} \color[HTML]{F1F1F1} 44.23 & {\cellcolor[HTML]{6AAED6}} \color[HTML]{F1F1F1} 73.65 \\
 & BLT & {\cellcolor[HTML]{F7FBFF}} \color[HTML]{000000} 0.05 & {\cellcolor[HTML]{F7FBFF}} \color[HTML]{000000} 18.21 & {\cellcolor[HTML]{F7FBFF}} \color[HTML]{000000} 13.68 & {\cellcolor[HTML]{F7FBFF}} \color[HTML]{000000} 64.46 \\
 & LayoutDM & {\cellcolor[HTML]{6AAED6}} \color[HTML]{F1F1F1} 8.55 & {\cellcolor[HTML]{ABD0E6}} \color[HTML]{000000} 26.96 & {\cellcolor[HTML]{ABD0E6}} \color[HTML]{000000} 38.06 & {\cellcolor[HTML]{ABD0E6}} \color[HTML]{000000} 72.44 \\
 & MaskGIT & {\cellcolor[HTML]{3787C0}} \color[HTML]{F1F1F1} 9.90 & {\cellcolor[HTML]{6AAED6}} \color[HTML]{F1F1F1} 30.99 & {\cellcolor[HTML]{3787C0}} \color[HTML]{F1F1F1} 48.12 & {\cellcolor[HTML]{3787C0}} \color[HTML]{F1F1F1} 74.04 \\
 & VQDiffusion & {\cellcolor[HTML]{ABD0E6}} \color[HTML]{000000} 8.28 & {\cellcolor[HTML]{D6E6F4}} \color[HTML]{000000} 26.05 & {\cellcolor[HTML]{D6E6F4}} \color[HTML]{000000} 37.73 & {\cellcolor[HTML]{D6E6F4}} \color[HTML]{000000} 72.22 \\
\bottomrule
\end{tabular}
\caption{Evaluation results on layout completion. The darker colors indicate the higher ranks. The measurement values are scaled by 100 for visibility.}
\label{tab:completion}
\end{table}

\newpage
\section{Visualizing Element Matching in \ourmeasure{}}
\begin{figure}[h!]
    \begin{minipage}[t]{0.47\linewidth}
        \includegraphics[clip,width=\linewidth]{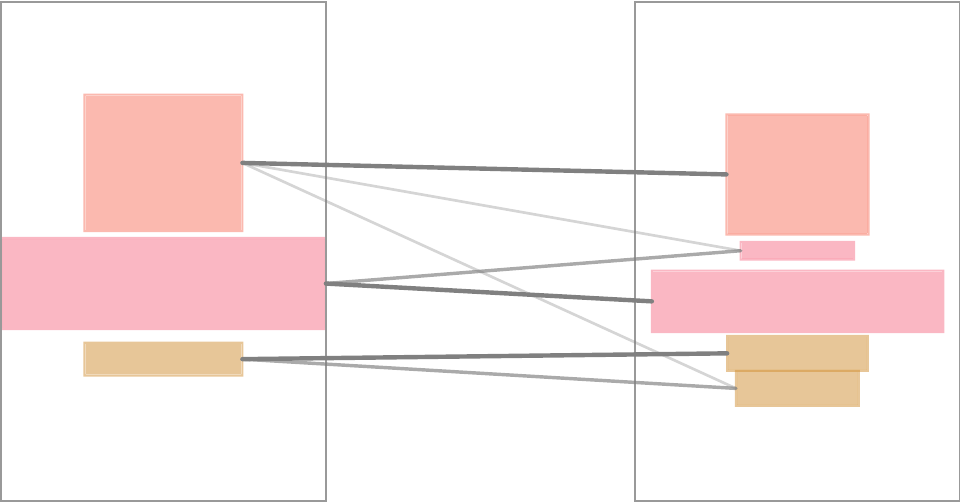}
    \end{minipage}
    \hspace{0.04\linewidth}
    \begin{minipage}[t]{0.47\linewidth}
        \includegraphics[clip,width=\linewidth]{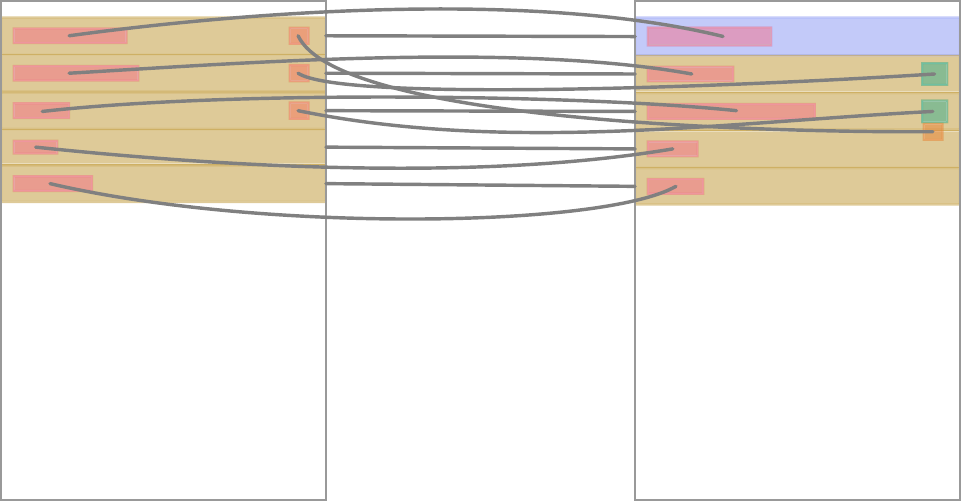}
    \end{minipage}
    \caption{Visualization of soft element matching in \ourmeasure{}. These edges, representing the optimal transportation $\gamma^\ast$, are thicker for matches with a higher weight.}
    \label{fig:elem_matching}
\end{figure}

Figure \cref{fig:elem_matching} illustrates the soft element matching in \ourmeasure{}. 
The gray edges represent the optimal transportation $\gamma^\ast_{i,j}$, with thicker edges indicating larger weights. 
The examples demonstrate that our measure exploits cross-category and many-to-many element matching.

\newpage
\section{Additional Retrieval Results}
We show additional retrieval results in the following figures.
For each query, top-5 retrieval results by \ourmeasure{}, DocSim, MeanIoU, and \deepsim{} are displayed.

\begin{figure}[h!]
    \centering
    \textsc{RICO}\\
    \includegraphics[clip,width=\linewidth]{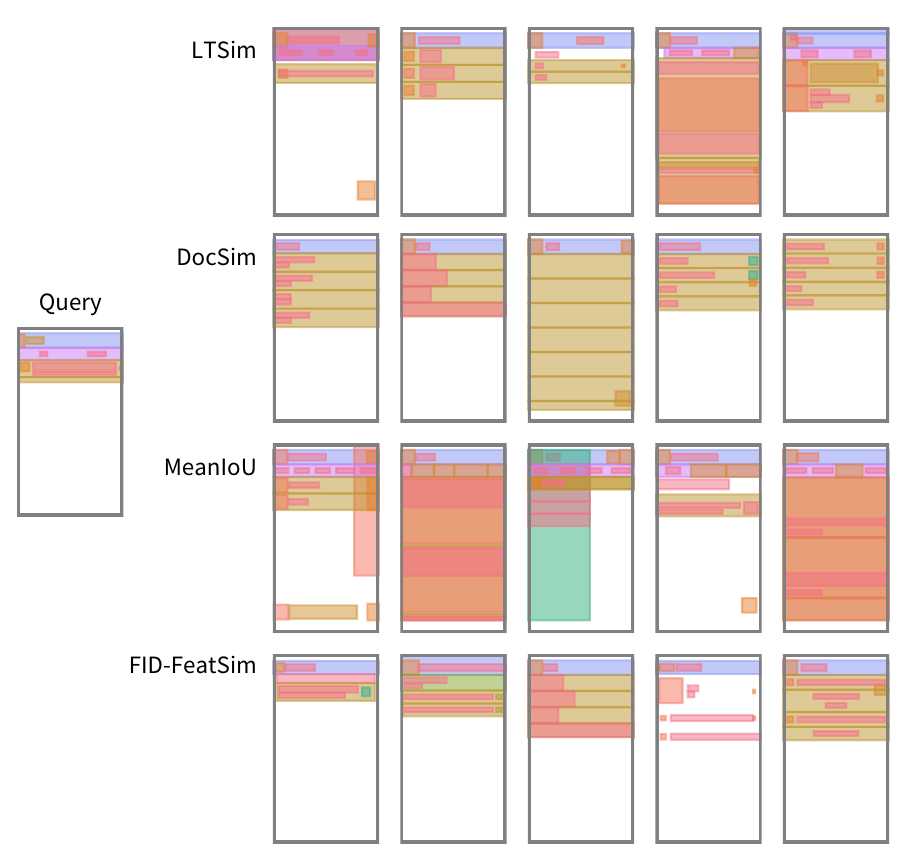}
    \caption{Top-5 retrieval results by \ourmeasure{}, DocSim, MeanIoU, \deepsim{}.}
    \label{fig:rico_516}
\end{figure}

\begin{figure}[h!]
    \centering
    \textsc{RICO}\\
    \includegraphics[clip,width=\linewidth]{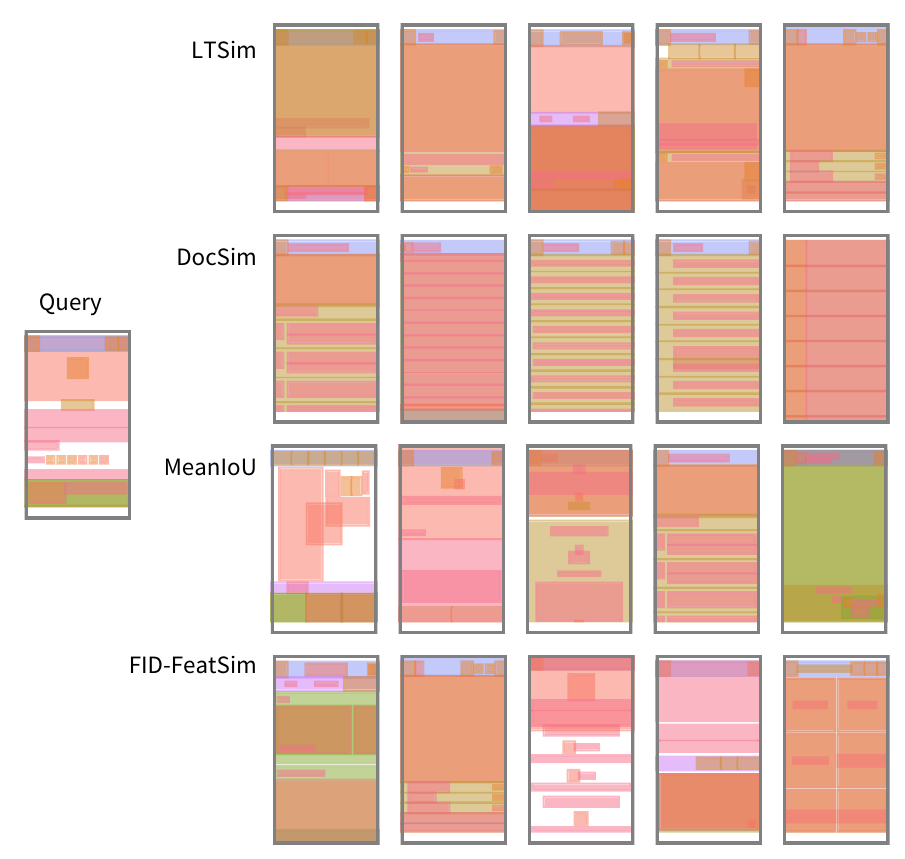}
\end{figure}

\begin{figure}[h!]
\centering
\textsc{RICO}\\
\includegraphics[clip,width=\linewidth]{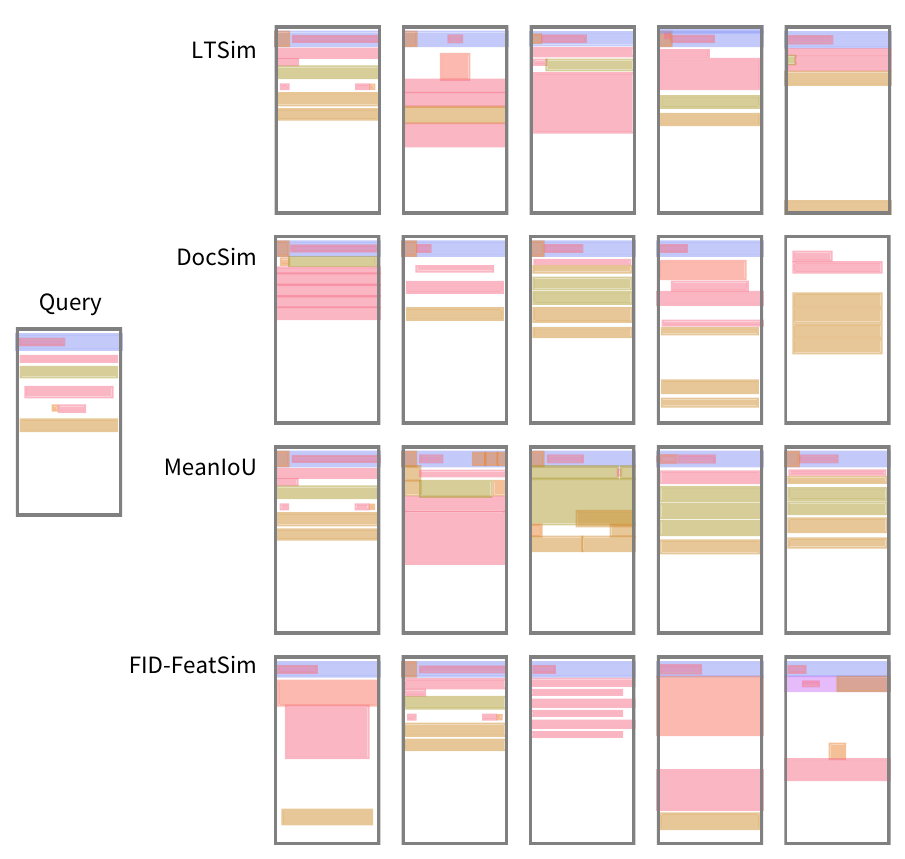}
\end{figure}

\begin{figure}[h!]
\centering
\textsc{RICO}\\
\includegraphics[clip,width=\linewidth]{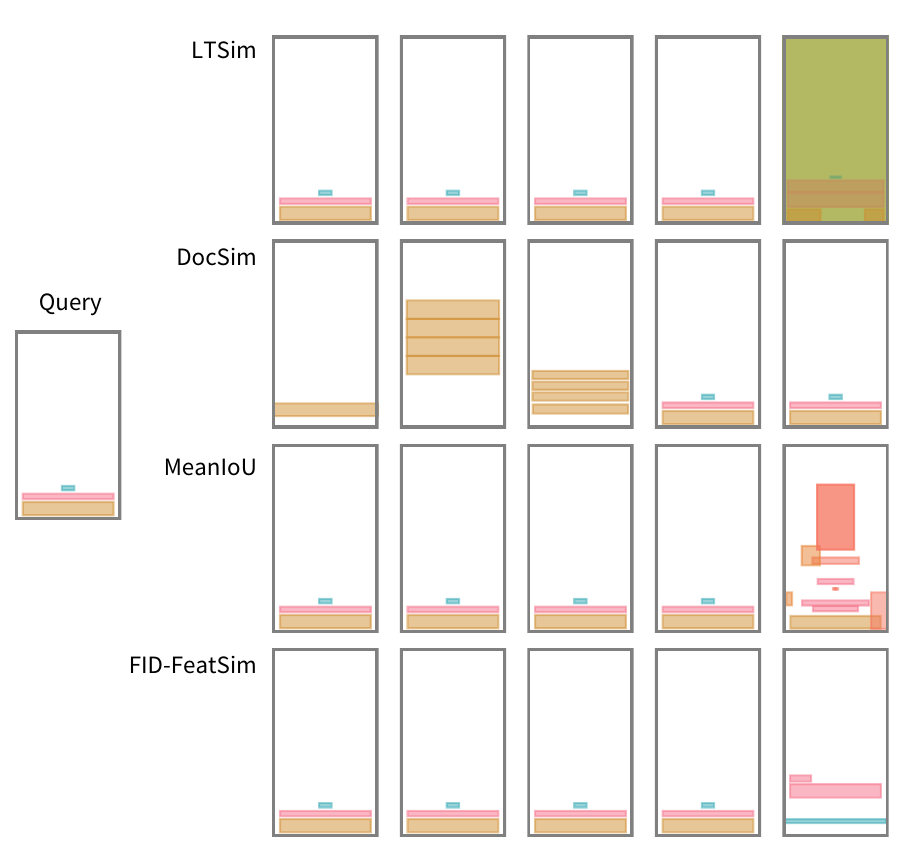}
\end{figure}

\begin{figure}[h!]
\centering
\textsc{PubLayNet}\\
\includegraphics[clip,width=\linewidth]{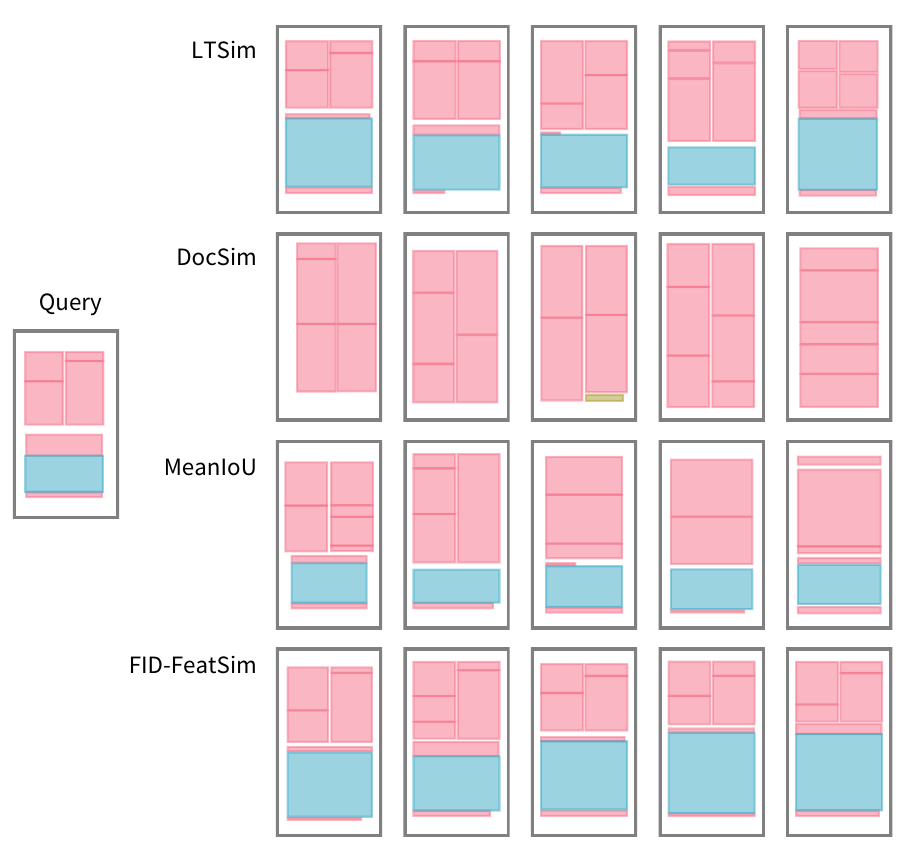}
\end{figure}

\begin{figure}[h!]
\centering
\textsc{PubLayNet}\\
\includegraphics[clip,width=\linewidth]{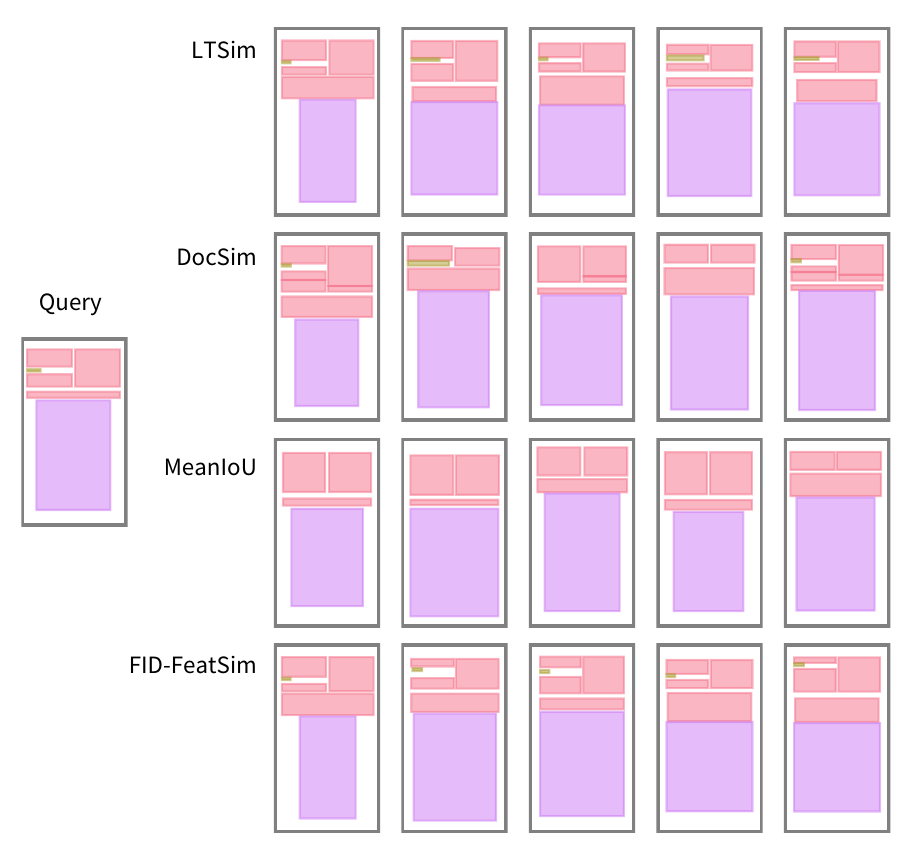}
\end{figure}

\begin{figure}[h!]
\centering
\textsc{PubLayNet}\\
\includegraphics[clip,width=\linewidth]{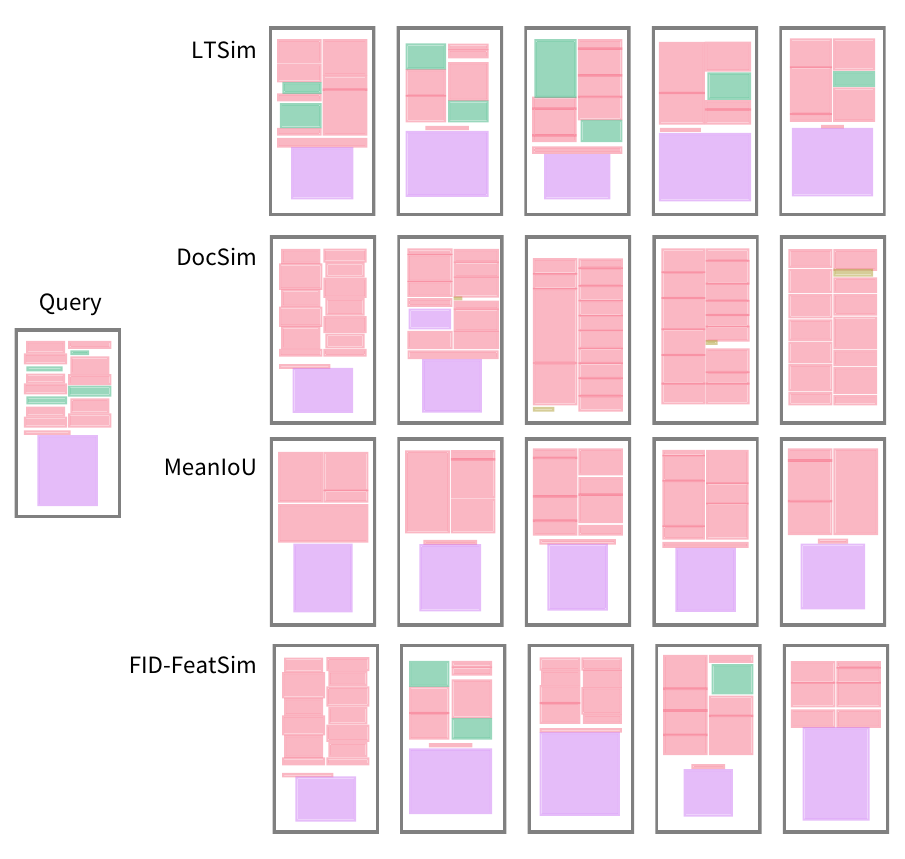}
\end{figure}

\begin{figure}[h!]
\centering
\textsc{PubLayNet}\\
\includegraphics[clip,width=\linewidth]{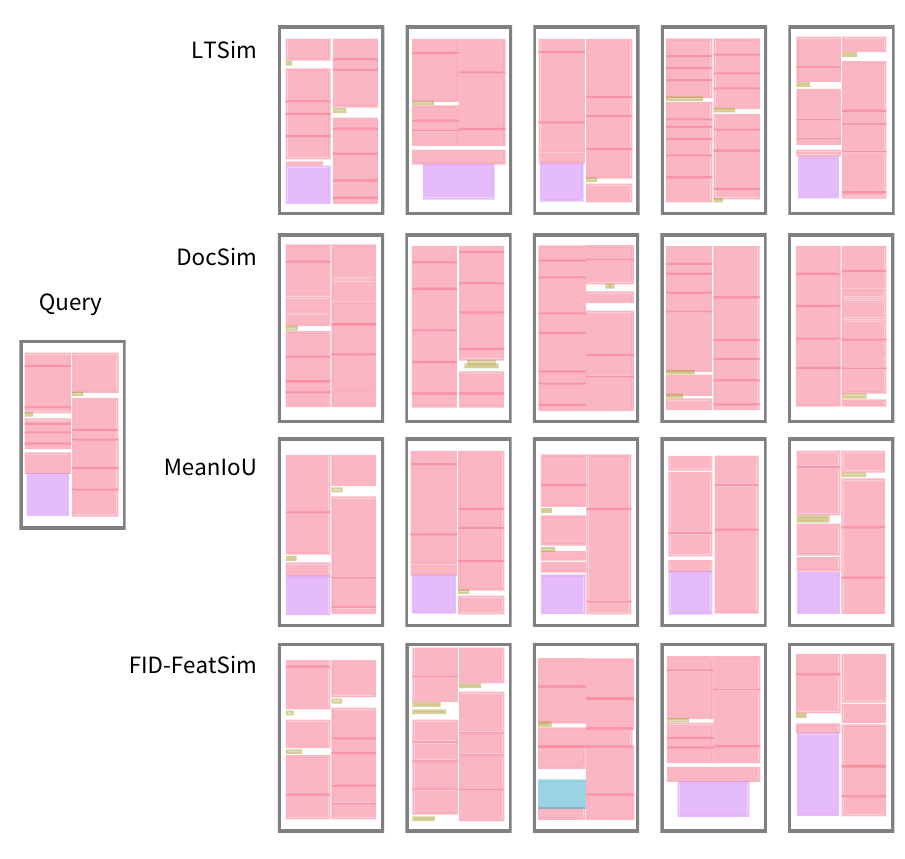}
\end{figure}

\clearpage
{
    \small
    \bibliographystyle{ieeenat_fullname}
    \bibliography{main}
}